\documentclass[lettersize,journal]{IEEEtran}
\usepackage{amsmath,amssymb,amsfonts}
\usepackage{algorithmic}
\usepackage{array}
\usepackage[caption=false,font=normalsize,labelfont=sf,textfont=sf]{subfig}
\usepackage{textcomp}
\usepackage{stfloats}
\usepackage{url}
\usepackage{verbatim}
\usepackage{graphicx}

\usepackage{cite}

\usepackage{gensymb}

\usepackage{multirow,multicol}
\usepackage{xcolor}

\usepackage{hyperref}   

\usepackage[capitalize]{cleveref}
\crefname{section}{Sec.}{Secs.}
\Crefname{section}{Section}{Sections}
\Crefname{table}{Table}{Tables}
\crefname{table}{Tab.}{Tabs.}
\Crefname{figure}{Figure}{Figures}
\crefname{figure}{Fig.}{Figs.}

\newcommand{\condensedpara}[1]{\vspace{0.25em}\noindent\textbf{#1}} 

\usepackage{booktabs}

\def\BibTeX{{\rm B\kern-.05em{\sc i\kern-.025em b}\kern-.08em
    T\kern-.1667em\lower.7ex\hbox{E}\kern-.125emX}}
\usepackage{balance}

\definecolor{LIME}{RGB}{  112,  173, 71}
\definecolor{CYAN}{RGB}{  91,  155, 213}

\begin{document}

\title{A Survey on Text-Driven 360-Degree Panorama Generation} 

\author{Hai Wang, Xiaoyu Xiang, Weihao Xia, and Jing-Hao Xue, \IEEEmembership{Senior Member, IEEE}
\thanks{H.~Wang, W.~Xia and J.-H.~Xue are with the Department of Statistical Science, University College London, London WC1E 6BT, U.K. (e-mail: hai.wang.22@ucl.ac.uk).}
\thanks{X.~Xiang is with the Core AI team at Meta Reality Labs, Menlo Park, CA 94025, USA.}
\thanks{Corresponding author: Hai Wang}
}


\maketitle

\begin{abstract}
The advent of text-driven 360-degree panorama generation, enabling the synthesis of 360-degree panoramic images directly from textual descriptions, marks a transformative advancement in immersive visual content creation. This innovation significantly simplifies the traditionally complex process of producing such content. Recent progress in text-to-image diffusion models has accelerated the rapid development in this emerging field. This survey presents a comprehensive review of text-driven 360-degree panorama generation, offering an in-depth analysis of state-of-the-art algorithms. We extend our analysis to two closely related domains: text-driven 360-degree 3D scene generation and text-driven 360-degree panoramic video generation. Furthermore, we critically examine current limitations and propose promising directions for future research. A curated project page with relevant resources and research papers is available at \href{https://littlewhitesea.github.io/Text-Driven-Pano-Gen/}{https://littlewhitesea.github.io/Text-Driven-Pano-Gen/}.
\end{abstract}


\section{Introduction}

Rapid growth of immersive technologies, such as virtual reality (VR) and augmented reality (AR), has dramatically increased the demand for high-quality panoramic visual content. Among such content, 360-degree panoramas are pivotal in delivering realistic and immersive experiences by capturing a complete spherical view of an environment. Traditionally, producing these panoramas requires specialized camera equipment and considerable technical expertise. However, recent advances in text-driven 360-degree panorama generation \cite{immerseGAN,diffpano,text2light,stitchdiffusion,diffusion360,panfusion} have introduced groundbreaking capabilities, enabling the synthesis of 360-degree panoramic images directly from textual descriptions. This innovation not only revolutionizes content creation over diverse domains \cite{interact360,virtualreality,gaming,dreamspace} including VR/AR applications, gaming, and virtual tours, but also serves as a foundational technology for new creative frontiers.

\begin{figure}[t!]
    \centering
    \includegraphics[width=0.5\textwidth]{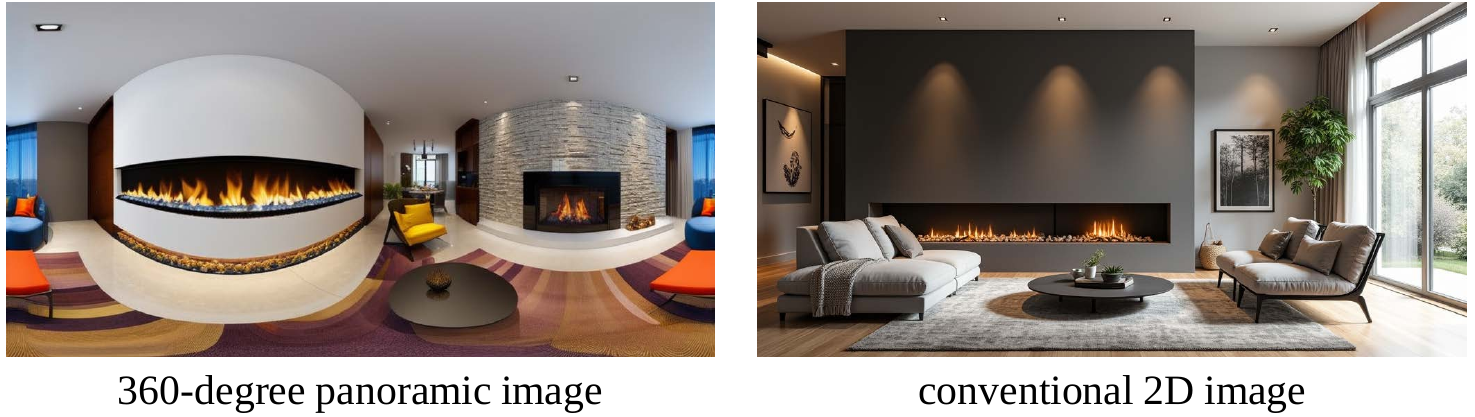}
    \caption{Visual comparison between a 360-degree panoramic image and a conventional 2D image.}
    \label{fig:teaser}
\end{figure}

Unlike conventional 2D images, 360-degree panoramic images, often represented through equirectangular projection \cite{360survey}, encompass the entire 360$\degree$$\times$180$\degree$ field of view, as shown in \cref{fig:teaser}. This distinctive format poses unique challenges for text-driven generation, requiring not only accurate image synthesis but also excellent preservation of geometric consistency and seamless visual coherence across the full $360\degree$ horizontal and $180\degree$ vertical extents.

The availability of large-scale paired image-text datasets has facilitated the development of text-to-image latent diffusion models (LDMs) \cite{latentdiffusion}, which excel at synthesizing high-quality, visually compelling images aligned with given text descriptions \cite{tcsvtt2i,t2i-survey,t2i-survey2,t2i-survey3,tcsvtsurvey}. Leveraging the powerful generative capabilities of pre-trained LDMs, researchers have developed methods specifically tailored to address the unique challenges of text-driven 360-degree panoramic image generation \cite{panodiff,stitchdiffusion,diffusion360,panfusion,aognet,spotdiffusion}. Although broader surveys on panoramic vision and 3D scene-generation~\cite{aisurvey,3dscenesurvey} briefly discuss some text-driven 360-degree panorama generation methods, they treat them only as peripheral topics.

To the best of our knowledge, a focused and systematic analysis devoted specifically to text-driven 360-degree panoramic image generation has not yet been presented. To address this gap, this paper presents a holistic survey and analysis of text-driven 360-degree panorama generation, its direct applications, and related emerging fields.

\begin{figure*}
    \centering
    \includegraphics[width=0.9\textwidth]{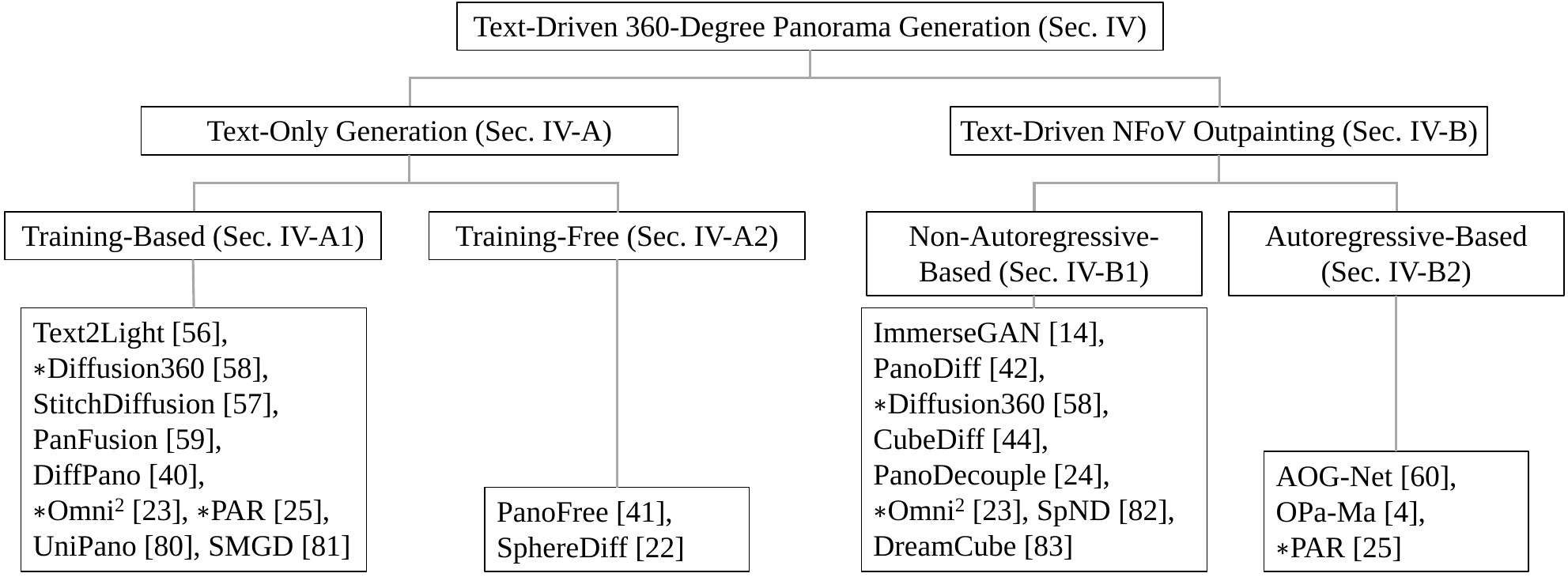}
    \caption{A systematic taxonomy proposed in this survey of text-driven 360-degree panorama generation methods. Methods marked with $*$ support multiple input modalities and therefore appear in more than one branch.}
    \label{fig:taxonomy}
\end{figure*}

\begin{figure*}
    \centering
    \includegraphics[width=0.9\textwidth]{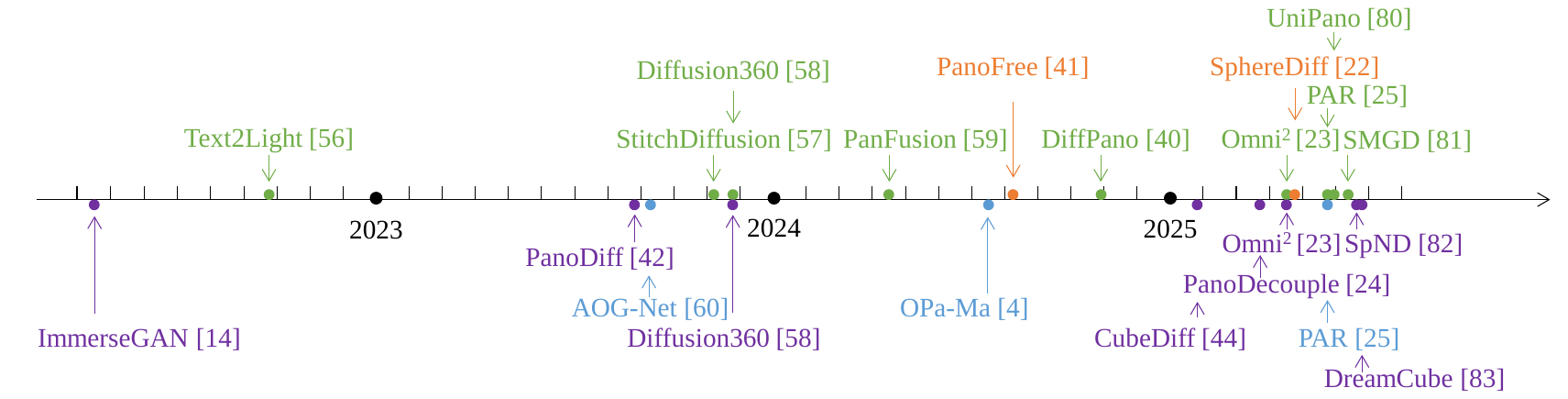}
    \caption{Chronological overview of text-driven 360-degree panorama generation approaches. Methods in \textcolor{LIME}{lime}, \textcolor{orange}{orange}, \textcolor{violet}{violet}, and \textcolor{CYAN}{cyan} are from \cref{subsubsec:training-based}, \cref{subsubsec:training-free}, \cref{subsubsec:nar-based}, and  \cref{subsubsec:ar-based}, respectively.}
    \label{fig:chronological}
\end{figure*}

This survey is structured as follows: First, we establish a foundational understanding of this field by introducing the principal representations of 360-degree panoramas, presenting prominent datasets commonly used in this area, and outlining key evaluation metrics employed to assess the quality and fidelity of generated panoramic content. Next, we review state-of-the-art (SOTA) methods for text-driven 360-degree panorama generation, categorizing them into two primary paradigms: (a) \textit{Text-Only Generation} and (b) \textit{Text-Driven Narrow Field-of-View (NFoV) Outpainting}. \cref{fig:taxonomy} and \cref{fig:chronological} provide a systematic taxonomy and a chronological overview of these SOTA methods, respectively. Following this, we explore two key emerging directions that are closely related to this field: (a) text-driven 360-degree 3D scene generation, which uses 360-degree panoramic images as an intermediate step; and (b) text-driven 360-degree panoramic video generation, a parallel and more complex task. Finally, we discuss the prevailing challenges in this developing field and propose potential directions for future research.

In short, this paper offers the first dedicated and comprehensive survey on text-driven 360-degree panoramic image generation, systematically reviewing its state-of-the-art techniques, key datasets and evaluation metrics. Furthermore, we explore its two closely related emerging directions: text-driven 360-degree 3D scene generation and text-driven 360-degree panoramic video synthesis. We also identify critical challenges and outline future research directions, aiming to offer a valuable resource to researchers and practitioners in this area.

\section{Related Work}

\subsection{Text-to-Image Diffusion Models}

Text-to-image (T2I) diffusion models \cite{glide,imagen,dall-e2,sdxl,latentdiffusion} have achieved remarkable progress in generating high-fidelity and photorealistic images from textual descriptions. These models have garnered widespread attention because of their intuitive text-based conditioning as a user-friendly interface for diverse image generation tasks.

T2I diffusion models can be broadly categorized into pixel-space and latent-space models. Pixel-space models, such as GLIDE \cite{glide} and Imagen \cite{imagen}, operate directly in the pixel space, producing visually impressive results at the expense of substantial computational resources, limiting their scalability. In contrast, latent diffusion models (LDMs) \cite{latentdiffusion} address these limitations by leveraging pre-trained autoencoders like VQGAN \cite{vqgan} to map images into a compact latent space, where the diffusion process is conducted. This reduces computational overhead while maintaining high-quality outputs, making LDMs a preferred framework for text-driven 360-degree panorama generation, as surveyed in this work.

\subsection{3D Scene Representation}

Efficient and accurate 3D scene representation is a critical challenge in computer graphics and vision. Traditional explicit representations, including point clouds, meshes, and voxel grids, often suffer from high memory requirements and struggle with complex topologies and unbounded scenes.

Neural implicit functions \cite{sdf,occupancy,3dgen}, which represent 3D scenes as continuous functions encoded within neural network parameters, offer a compact and flexible paradigm for scene representation. Notably, Neural Radiance Fields (NeRFs) \cite{nerf} stand out for their ability to achieve high-quality novel view synthesis. However, NeRF's reliance on dense volumetric sampling along camera rays results in slow training, hindering its practicability.

Recently, 3D Gaussian Splatting (3DGS) \cite{3dgs} has emerged as an efficient alternative to 3D scene representation. By combining an explicit representation of 3D Gaussians with a highly efficient differentiable rasterization pipeline, 3DGS facilitates rapid scene reconstruction and rendering. This advancement has opened up new possibilities, including recent explorations in text-to-3D 360-degree scene synthesis~\cite{scenedreamer360,dreamscene360}, which leverage text-driven 360-degree panorama generation techniques.

\section{Preliminaries}

\subsection{Representations of 360-Degree Panoramas}
\label{subsec:representations}

\begin{figure}[t!]
    \centering
    \includegraphics[width=0.48\textwidth]{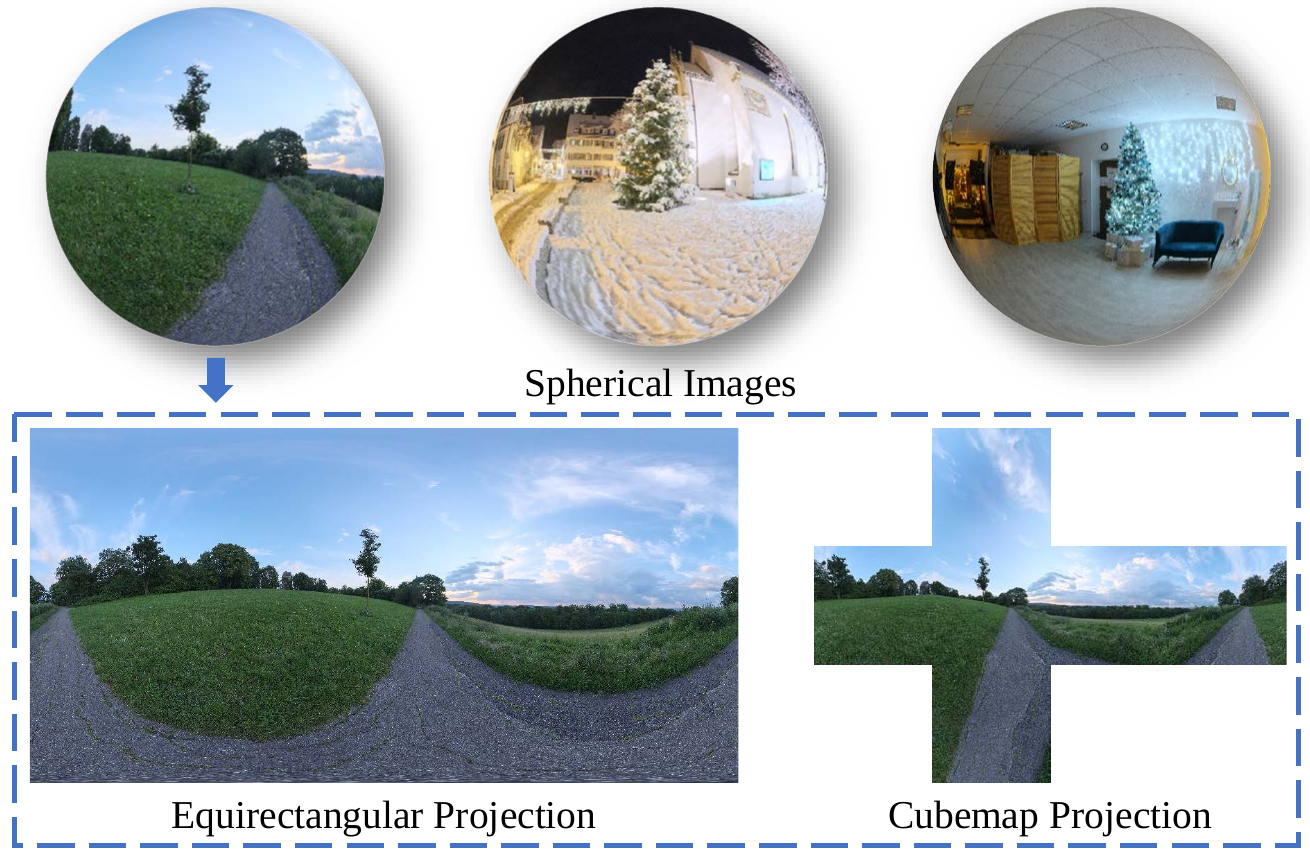}
    \caption{Visual comparison between the equirectangular and cubemap projections of spherical images (360-degree panoramic images).}
    \label{fig:erp_cmp}
\end{figure}

The representation of 360-degree panoramic content poses a fundamental challenge: How to accurately map spherical visual information onto a two-dimensional plane? To address this, a variety of projection methodologies \cite{360survey,aisurvey} have been developed, each with distinct advantages and trade-offs. Below, we first outline two widely used formats for 360-degree panorama representation: Equirectangular Projection (ERP) and Cubemap Projection (CMP), as illustrated in \cref{fig:erp_cmp}, and then introduce Multi-Perspective Projection (MPP), a category we define for the classification purposes of this survey.

\subsubsection{Equirectangular Projection (ERP)} 

As the most prevalent representation format for 360-degree panoramas, ERP establishes a direct mapping between spherical and planar coordinates: longitude corresponds to the horizontal axis, spanning the full 360$\degree$ range, while latitude maps to the vertical axis, covering 180$\degree$ from -90$\degree$ (south pole) to +90$\degree$ (north pole). ERP's simplicity and compatibility with web viewers and VR headsets make it the preferred choice for numerous applications. Additionally, its representation as a single, continuous image allows the direct application of image manipulation techniques, such as text-driven 360-degree panorama-to-panorama translation \cite{360pant}. Despite these advantages, ERP introduces pronounced geometric distortions, particularly at the polar regions, where the visual content appears stretched. Furthermore, the texel density of a spherical image in ERP is non-uniformly distributed: it is comparatively lower in the equatorial regions and markedly higher towards the poles. This inhomogeneity can be particularly problematic in scenarios where critical visual information is predominantly located away from the poles, leading to inefficient utilization of image resolution for regions of interest. In this survey, unless explicitly stated otherwise, 360-degree panoramas are represented using ERP.

\subsubsection{Cubemap Projection (CMP)} 

CMP offers an alternative representation that mitigates the distortions inherent in the ERP format, particularly at the poles. In CMP, the spherical image is projected onto the six faces of a cube, with each face representing a 90$\degree$$\times$90$\degree$ field of view. This division significantly reduces geometric distortions, making CMP more compatible with diffusion priors from text-to-image diffusion models trained on standard perspective images \cite{cubediff}. However, CMP introduces several challenges: (1) it increases the complexity of image manipulation compared to the single-image format of ERP; (2) it may necessitate additional conversion for compatibility with platforms or viewers that primarily support ERP. Despite these practical challenges, CMP is well-suited for applications that demand reduced distortion and higher fidelity. The width and height of an ERP image are four and two times the side length of the corresponding CMP, respectively, reflecting the geometric relationship between the two formats.

\subsubsection{Multi-Perspective Projection (MPP)}

MPP is defined as the use of multiple individual perspective images to collectively represent a 360-degree panoramic view, where these images may or may not overlap. This category is characterized by configurations that deviate from the standard six-face, 90$\degree$$\times$90$\degree$ field of view CMP.

The advantages and limitations of these representation formats are further analyzed in \cref{subsubsec:choice_of_representation}.

\begin{table}[t]
\vspace{-3mm}
\small
    \caption{Summary of Popular Datasets Used for Text-Driven 360-Degree Panorama Generation: Includes categories, publication year, sample size, resolution (Res.), and license. Categories are indoor (I), outdoor (O), or hybrid (I, O). Datasets marked with $\star$ are sourced from public websites. 
    }
    \label{tab:dataset}
\centering
\begin{tabular}{lcrcl}
\toprule
Dataset (Category) & Year & \#Samples & Res.  & License \\ 
\midrule
SUN360 (I, O)    & 2012                   & 67,583       &  9K & Custom   \\ 
Matterport3D (I)     & 2017                 & 10,800       & 2K    & Custom         \\
Laval Indoor (I)      & 2017                 & 2,233       & 7K   & Custom   \\
Laval Outdoor (O)      & 2019                 & 205       & 7K   & Custom  \\
Structured3D (I)      & 2020                  & 196,515       & 1K   & Custom \\
Pano360 (I, O)      & 2021                & 35,000       & 8K   & Custom   \\
$\star$Polyhaven (I, O)      & 2025                  & 786       & 8K     & CC0       \\
$\star$Humus (I, O)      & 2025                  & 139       & 8K       & CC BY 3.0     \\
\bottomrule
\end{tabular}
\vspace{-3mm}
\end{table}

\subsection{Datasets}
\label{subsec:datasets}

360-degree panoramic image generation from text prompts presents unique challenges due to the complete 360$\degree$$\times$180$\degree$ field of view that these images encompass. Text-to-image diffusion models \cite{imagen,sdxl,tcsvtt2i2,latentdiffusion}, predominantly trained on perspective images with a narrower field of view, often struggle to synthesize high-quality 360-degree panoramas. To address this, several specialized datasets have been developed to facilitate research in this domain. \cref{tab:dataset} summarizes these datasets, with further details provided below.

\condensedpara{SUN360}~\cite{sun360} is a comprehensive database comprising 67,583 high-resolution (9104$\times$4552) panoramic images sourced from the Internet. Each image covers a 360$\degree$$\times$180$\degree$ field of view in ERP format and is manually categorized into 80 distinct classes. Originally created for scene viewpoint recognition, SUN360 now serves as a valuable resource for a wide range of computer vision, computer graphics, and related research areas.

\condensedpara{Matterport3D}~\cite{matterport3d} offers 10,800 indoor 360-degree panoramic images with corresponding depth maps, all at a resolution of 2048$\times$1024 pixels. These panoramas are derived from 194,400 RGB-D images of 90 buildings, making it a rich dataset for studying indoor environments.

\condensedpara{Laval Indoor}~\cite{lavalindoor} consists of 2,233 high-dynamic-range, high-resolution (7768$\times$3884) 360-degree panoramic images, specifically curated for the study of extensive indoor scenes, such as factories, apartments, and houses.

\condensedpara{Laval Outdoor}~\cite{lavaloutdoor} complements its indoor counterpart, offering 205 high-dynamic-range, high-resolution (7768$\times$3884) 360-degree panoramic images that capture diverse outdoor environments, including urban and natural scenes.

\condensedpara{Structured3D}~\cite{structured3d} contains 196,515 360-degree panoramas with varying configurations and lighting conditions, representing 21,835 distinct rooms. Rendered at a resolution of 1024$\times$512 from 3D scenes of original house design files, Structured3D is ideal for research on structured 3D modeling and understanding.

\condensedpara{Pano360}~\cite{pano360} contains 35,000 360-degree panoramic images with a resolution of 8192$\times$4096. Of these, 34,000 are sourced from Flickr, with the remainder rendered from photorealistic 3D scenes. Pano360 was originally proposed for training camera calibration networks.

\condensedpara{Polyhaven}~\cite{polyhaven} contributes 786 real-world high-resolution (8192$\times$4096) 360-degree panoramas encompassing a variety of indoor and outdoor scenes.

\condensedpara{Humus}~\cite{humus} includes 139 real-world 360-degree panoramas represented using cubemap projection, with each face having a resolution of 2048$\times$2048 pixels. This dataset includes indoor and outdoor environments.

\subsection{Evaluation Metrics}
\label{subsec:metrics}

A rigorous evaluation of text-driven 360-degree panorama generation methods typically requires to combine (a)~\textit{universal} and (b)~\textit{panorama-specific} metrics. The universal metrics, comprising Fréchet Inception Distance (FID), Kernel Inception Distance (KID), Inception Score (IS), and CLIP Score~(CS), are widely applicable to both perspective and panoramic images.

\condensedpara{FID}~\cite{fid} measures the distance between feature distributions of generated and real images using a pre-trained Inception-v3 network~\cite{inception}. Lower FID scores indicate better perceptual quality and closer alignment with the real image distribution.

\condensedpara{KID}~\cite{kid} measures the difference between real and generated image distributions by computing the maximum mean discrepancy of their features extracted from Inception-v3 \cite{inception}. Similar to FID, lower KID values indicate better image quality.

\condensedpara{IS}~\cite{is} measures both the quality and diversity of generated images by leveraging Inception-v3 \cite{inception}. It calculates the KL divergence between the conditional class distribution of generated images and the marginal distribution over all generated samples. Higher IS suggests better visual quality and diversity.

\condensedpara{CS}~\cite{clip} evaluates consistency between text prompts and generated images using the CLIP model~\cite{clip}. It calculates the cosine similarity between the text embedding of the prompt and the visual embedding of the generated image. A higher CS reflects stronger text-image alignment and semantic coherence.

\begin{figure*}
    \centering
    \includegraphics[width=\textwidth]{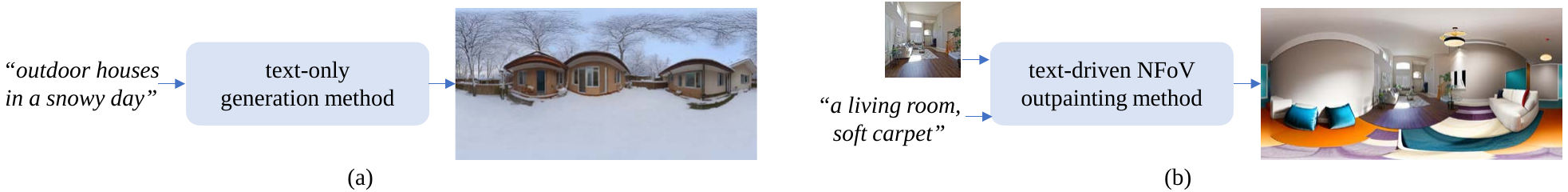}
    \caption{Paradigms for Text-Driven 360-Degree Panorama Generation. (a) Text-Only Generation synthesizes 360-degree panoramas from textual descriptions only. (b) Text-Driven NFoV Outpainting uses prompts and initial NFoV images as input to generate 360-degree panoramic images.}
    \label{fig:demo}
\end{figure*}

\begin{table*}[t]
\small
    \caption{Summary of Text-Only Generation Methods. `LDM-B' indicates whether the method is based on latent diffusion models. `TF' specifies if it is training-free. `Code' denotes whether both the source code and the pre-trained model checkpoint are publicly accessible. `N/A' means not applicable.}
    \label{tab:only}
    \centering
    \begin{tabular}{llccccc}
        \toprule
        Method & Publication & LDM-B & TF & Code & Training Datasets & Representation \\
        \midrule
        \multirow{2}{*}{Text2Light \cite{text2light}} & \multirow{2}{*}{TOG 2022} & \multirow{2}{*}{$\times$} & \multirow{2}{*}{$\times$} & \multirow{2}{*}{$\checkmark$} & Polyhaven \cite{polyhaven}, Laval Indoor \cite{lavalindoor}, Laval Outdoor \cite{lavaloutdoor} & \multirow{2}{*}{ERP} \\
        & & & & & iHDRI \cite{ihdri}, HDRI Skies \cite{hdri-skies}, HDRMaps \cite{hdrmaps} & \\
        Diffusion360 \cite{diffusion360} & arxiv 2023 & $\checkmark$    & $\times$ & $\checkmark$ & SUN360 \cite{sun360} & ERP \\
        StitchDiffusion \cite{stitchdiffusion} & WACV 2024 & $\checkmark$   & $\times$ & $\checkmark$ & Polyhaven \cite{polyhaven} & ERP \\
        PanFusion \cite{panfusion}    & CVPR 2024 & $\checkmark$    & $\times$ & $\checkmark$  & Matterport3D \cite{matterport3d} & ERP \\
        PanoFree \cite{panofree}    & ECCV 2024 & $\checkmark$    & $\checkmark$ & $\times$ & N/A & MPP \\
        DiffPano \cite{diffpano}    & NeurIPS 2024 & $\checkmark$     & $\times$ & $\times$ & Habitat Matterport 3D \cite{HM3D} & ERP \\
        $\text{Omni}^{2}$ \cite{omni2} & ACM MM 2025 & $\times$    & $\times$ & $\times$ & SUN360 \cite{sun360}, Structured3D \cite{structured3d}  & MPP \\
        SphereDiff \cite{spherediff} & arxiv 2025 & $\checkmark$    & $\checkmark$ & $\times$ & N/A & Spherical \\
        PAR \cite{par} & arxiv 2025 & $\times$    & $\times$ & $\checkmark$ & Matterport3D \cite{matterport3d} & ERP \\
        UniPano \cite{unipano} & ICCV 2025 & $\checkmark$    & $\times$ & $\times$ & Matterport3D \cite{matterport3d} & ERP \\
        SMGD \cite{smgd}    & CVPR 2025 & $\checkmark$    & $\times$ & $\checkmark$  & Matterport3D \cite{matterport3d} & Spherical \\
        \bottomrule
    \end{tabular}
\end{table*}

However, these universal metrics have significant limitations when applied to 360-degree panoramic content. The core issue lies in their underlying encoders (e.g., Inception-v3, CLIP's ViT), which were trained primarily on large datasets of standard perspective images. These encoders are not trained to account for the geometric distortions inherent in equirectangular projections. Consequently, FID, KID, and IS may penalize geometrically correct panoramic features as artifacts, leading to an inaccurate assessment of perceptual quality. Similarly, the CS may fail to accurately measure text-image alignment, as its encoder might misinterpret distorted objects or spatial relationships within the spherical scene. This gap highlights the inadequacy of universal metrics for capturing properties unique to 360-degree panoramas, such as seamlessness and global geometric fidelity. To address these shortcomings, several panorama-specific metrics have been proposed, including Fréchet Auto-Encoder Distance (FAED), Omnidirectional FID (OmniFID), Discontinuity Score (DS), and Distortion-perception FID (Distort-FID).

\condensedpara{FAED}~\cite{faed} computes Fréchet distances between features extracted from generated and real panoramas. Unlike FID, it employs an autoencoder~\cite{autoencoder} specifically trained on 360-degree panoramic images. Lower FAED scores reflect better perceptual and geometric quality tailored to the unique panoramic properties.

\condensedpara{OmniFID}~\cite{omnifid} adapts FID to specifically evaluate 360-degree panoramas. It converts equirectangular panoramas into cubemap representations and calculates FID across three disjoint subsets of cubemap faces, averaging the results. Lower OmniFID scores indicate higher geometric fidelity in 360-degree panorama generation.

\condensedpara{DS}~\cite{omnifid} measures the seam alignment across the borders of generated panoramas by applying a kernel-based edge detection algorithm. A lower DS corresponds to fewer visible seam artifacts, indicating better perceived consistencies across the seam.

\condensedpara{Distort-FID}~\cite{panodecouple} measures the distance between feature distributions of generated and real images based on a distortion-perception CLIP~\cite{panodecouple} that is fine-tuned on 360-degree panoramic images. Lower Distort-FID scores reflect better distortion accuracy of the generated 360-degree panoramic images.

Despite these advances, a critical gap remains. While metrics like OmniFID improve the measurement of geometric fidelity by mitigating ERP distortions and DS checks for visual seamlessness, they do not holistically evaluate the global semantic and structural coherence of a full 360-degree panoramic scene. Developing new metrics that capture this global consistency is therefore an important direction for future research.

We provide a comprehensive comparison of state-of-the-art methods, introduced in the following section, using the outlined metrics in~\cref{subsec:comparison}.

\section{State-of-the-Art Methods for Images}
\label{sec:methods}

Existing text-driven 360-degree panorama generation methods can be broadly categorized into two paradigms according to input modalities: (a)~\textit{Text-Only Generation} aims to synthesize 360-degree panoramas from textual prompts only, while (b) \textit{Text-Driven Narrow Field-of-View (NFoV) Outpainting} leverages both textual descriptions and initial NFoV images to guide the generation process, offering enhanced user control. \cref{fig:demo} provides an intuitive illustration for both paradigms. We detail the literature for both as follows.

\subsection{Text-Only Generation} 
\label{subsec:text_only}

This paradigm focuses on synthesizing 360-degree panoramas from textual descriptions only. \cref{tab:only} provides a comparative overview of representative text-only methods. These methods can be broadly divided into training-based and training-free approaches.

\subsubsection{Training-Based}
\label{subsubsec:training-based}

Text2Light~\cite{text2light}, an early notable effort, explores a hierarchical text-driven framework, using VQGAN~\cite{vqgan} and CLIP~\cite{clip} to address this challenge based on training data aggregated from multiple sources, such as Polyhaven~\cite{polyhaven}, Laval Indoor~\cite{lavalindoor} and Laval Outdoor~\cite{lavaloutdoor}. Recently, the advent of latent diffusion models (LDMs)~\cite{latentdiffusion} for text-to-image synthesis marks a significant advancement, enabling more sophisticated 360-degree panorama generation techniques. LDMs are typically trained on vast datasets consisting of standard perspective images with a limited field of view and corresponding textual descriptions. 
Despite demonstrating robust capabilities in generating perspective images from text prompts, these models face significant difficulties when creating 360-degree panoramas with a complete 360$\degree$$\times$180$\degree$ field of view, which differ substantially from traditional perspective images.

To adapt pre-trained LDMs for 360-degree panorama synthesis, a common strategy is to fine-tune these models with specialized 360-degree panorama datasets. Diffusion360~\cite{diffusion360} exemplifies this approach by leveraging the DreamBooth technique~\cite{dreambooth} to fine-tune a pre-trained LDM~\cite{latentdiffusion} on SUN360~\cite{sun360}. To ensure geometric consistency of boundaries, Diffusion360 uses a circular blending strategy during both the denoising process and the VAE~\cite{vae} decoding stage, effectively reducing seam artifacts. In addition, it introduces a super-resolution module to enable the generation of high-resolution (6114$\times$3072) 360-degree panoramas. While this full fine-tuning approach adopted by Diffusion360 effectively embeds panoramic geometry into the model, its primary trade-off is the high computational cost and risk of quality drop by altering the model's original generative priors.

In contrast, LoRA~\cite{lora} has recently gained attention as a parameter-efficient fine-tuning method. LoRA works by injecting trainable low-rank matrices into the pre-trained model's weights, allowing for rapid adaptation to new tasks with minimal additional parameters. For example, StitchDiffusion~\cite{stitchdiffusion} employs LoRA to fine-tune a pre-trained LDM on a dataset of 120 paired 360-degree images (sourced from Polyhaven~\cite{polyhaven}) and corresponding textual descriptions generated using BLIP~\cite{blip}. Its key contribution is formulating panorama generation as a latent-space stitching problem, using a MultiDiffusion-based~\cite{multidiffusion} method to enforce boundary continuity. However, the geometry fidelity of 360-degree panoramas generated by StitchDiffusion is relatively low due to the small fine-tuning dataset.

Other works \cite{diffpano,panfusion,unipano,latentlabs360} have similarly adopted the LoRA fine-tuning technique. PanFusion \cite{panfusion} contributes a novel dual-branch architecture, trained on the Matterport3D~\cite{matterport3d} dataset, with separate LoRA layers to integrate both global panoramic and local perspective views. It introduces an equirectangular-perspective projection attention module to facilitate information exchange between the two branches, aiming to alleviate visual inconsistencies in the generated 360-degree panoramas. However, PanFusion's output often exhibits blurriness at the top and bottom regions, due to its training dataset. To avoid this issue, DiffPano~\cite{diffpano} uses the Habitat Matterport 3D dataset~\cite{HM3D} to produce multi-view consistent 360-degree panoramas with clearer top and bottom details. For generating more precise textual descriptions, DiffPano adopts BLIP2 \cite{blip2} and Llama2 \cite{llama2} sequentially, resulting in a panoramic video-text dataset. Based on this dataset, it fine-tunes a pre-trained LDM~\cite{latentdiffusion} using LoRA for single-view text-driven 360-degree panorama generation. Furthermore, to enable multi-view 360-degree panorama generation based on text prompts and camera viewpoints, DiffPano introduces a spherical epipolar-aware attention module, a key innovation for enforcing multi-view geometric consistency. Based on the observation that value and output weight matrices are crucial during the LoRA-based fine-tuning of cross-attention blocks within pre-trained LDMs for 360-degree panoramic image generation, UniPano~\cite{unipano} proposes a uni-branch framework for panorama generation. This framework exclusively fine-tunes these specific matrices using LoRA, while keeping the original query and key weight matrices in the cross-attention blocks frozen.

Diverging from fine-tuning the commonly used pre-trained LDMs~\cite{latentdiffusion} (as in PanFusion~\cite{panfusion} and DiffPano~\cite{diffpano}), several studies have explored alternative pre-trained models for synthesizing 360-degree panoramic images from textual descriptions. Inspired by OmniGen~\cite{omnigen}, $\text{Omni}^{2}$~\cite{omni2} adopts a diffusion model consisting of a VAE~\cite{vae} and a pre-trained Transformer \cite{transformer}. To adapt the pre-trained Transformer for synthesizing 360-degree panoramic images, the LoRA fine-tuning technique is employed. Instead of processing the entire 360-degree panoramic image at once, $\text{Omni}^{2}$ generates six overlapping viewports. In the inference phase, these synthesized perspective images are integrated to reconstruct 360-degree panoramic images. Addressing the challenge that spatial distortions in ERP 360-degree panoramic images violate the identically and independently distributed (i.i.d.) Gaussian noise assumption inherent in many diffusion models, PAR~\cite{par}, inspired by masked autoregressive modeling (MAR)~\cite{mar}, proposes an autoregressive modeling approach for text-based 360-degree panoramic image generation. This method is not constrained by the i.i.d. assumption. Specifically, PAR fine-tunes a pre-trained autoregressive model~\cite{nova} on the Matterport3D dataset~\cite{matterport3d} and develops a dual-space circular padding technique to mitigate boundary discontinuities. 

Most aforementioned approaches rely on ERP representations, which struggle to adequately deal with the inherent spherical distortions. To mitigate these distortions and maintain global geometric coherence, SMGD~\cite{smgd} proposes the use of spherical manifold convolution within a spherical manifold U-Net combined with VQGAN~\cite{vqgan}, enabling more accurate synthesis of 360-degree panoramic images, but the primary trade-off is reduced transferability of pre-trained diffusion priors, since the specialized spherical convolutions are not directly compatible with standard text-to-image diffusion architectures~\cite{latentdiffusion,sdxl}.

\subsubsection{Training-Free}
\label{subsubsec:training-free}

In contrast to training-based approaches, training-free methods avoid any model fine-tuning and instead repurpose powerful pre-trained text-to-image diffusion backbones. PanoFree \cite{panofree} pioneered a tuning-free multi-view image generation framework based on a pre-trained LDM \cite{latentdiffusion}. Guided by textual descriptions, PanoFree leverages iterative warping and inpainting steps to produce multi-view perspective images, which are subsequently stitched into 360-degree panoramas, thus avoiding the need for specialized 360-degree panorama datasets. While this avoids the need for specialized training data, its multi-step process can be slow and risks accumulating errors that harm global consistency. Unlike PanoFree, which operates on perspective latent representations, the recent SphereDiff~\cite{spherediff} contributes a more theoretically grounded approach by extending the MultiDiffusion~\cite{multidiffusion} framework to a spherical latent space. To mitigate minor distortions arising from the spherical-to-perspective projection during its process, SphereDiff further incorporates a distortion-aware weighted averaging method. Although these approaches inherit the rich prior knowledge of large text-to-image models and require no additional training or 360-degree panoramic data, their patch-based synthesis mechanisms can lead to global inconsistencies and comparatively longer inference times than training-based models. Future work may explore stronger global guidance and more efficient inference designs to overcome these limitations.

\begin{table*}
\small
    \caption{Summary of Text-Driven NFoV Outpainting Methods. 
    For explanations of the `LDM-B', `TF' and `Code' columns, see  \cref{tab:only}.
    }
    \label{tab:outpainting}
    \centering
    \begin{tabular}{llccccc}
        \toprule
        Method & Publication & LDM-B & TF & Code & Training Datasets & Representation \\
        \midrule
        ImmerseGAN \cite{immerseGAN} & 3DV 2022 & $\times$     & $\times$ & $\times$ & 360Cities \cite{360cities} & ERP \\
        PanoDiff \cite{panodiff} & MM 2023 & $\checkmark$    & $\times$ & $\checkmark$ & SUN360 \cite{sun360} & ERP \\
        Diffusion360 \cite{diffusion360} & arxiv 2023 & $\checkmark$    & $\times$ & $\checkmark$ & SUN360 \cite{sun360} & ERP \\
        AOG-Net \cite{aognet} & AAAI 2024 & $\checkmark$    & $\times$ & $\times$ & Laval Indoor \cite{lavalindoor}, Laval Outdoor \cite{lavaloutdoor} & ERP\\
        OPa-Ma \cite{opa-ma}    & arxiv 2024 & $\checkmark$     & $\times$ & $\times$ & Laval Indoor \cite{lavalindoor}, Laval Outdoor \cite{lavaloutdoor} & ERP\\
        \multirow{2}{*}{CubeDiff \cite{cubediff}}    & \multirow{2}{*}{ICLR 2025} & \multirow{2}{*}{$\checkmark$}   & \multirow{2}{*}{$\times$} & \multirow{2}{*}{$\times$} & Polyhaven \cite{polyhaven}, Humus \cite{humus}, Structured3D \cite{structured3d} &  \multirow{2}{*}{CMP} \\
        & & & & & Pano360 \cite{pano360}  & \\
        PanoDecouple \cite{panodecouple} & CVPR 2025 & $\checkmark$    & $\times$ & $\times$ & SUN360 \cite{sun360} & ERP \\
        $\text{Omni}^{2}$ \cite{omni2} & ACM MM 2025 & $\times$    & $\times$ & $\times$ & SUN360 \cite{sun360}, Structured3D \cite{structured3d} & MPP \\
        PAR \cite{par} & arxiv 2025 & $\times$    & $\times$ & $\times$ & Matterport3D \cite{matterport3d} & ERP \\
        SpND \cite{spnd} & ICML 2025 & $\checkmark$    & $\times$ & $\checkmark$ & Matterport3D \cite{matterport3d}, Structured3D \cite{structured3d}  & ERP \\
        DreamCube \cite{dreamcube} & ICCV 2025 & $\checkmark$    & $\times$ & $\checkmark$ &  Structured3D \cite{structured3d}  & CMP \\
        \bottomrule
    \end{tabular}
\end{table*}

\subsection{Text-Driven NFoV Outpainting} 
\label{subsec:outpainting}

This paradigm enhances user control by conditioning the 360-degree panorama generation process on both textual prompts and initial NFoV images. The NFoV image, representing a limited portion of the scene, serves as a visual starting point, which the generative model subsequently expands into a complete 360-degree panoramic image guided by the textual description. \cref{tab:outpainting} offers a summary of representative text-driven NFoV outpainting approaches. These approaches can be broadly classified into non-autoregressive-based (NAR-based) and autoregressive-based (AR-based) methods according to their underlying frameworks. 

\subsubsection{NAR-Based}
\label{subsubsec:nar-based}

An early attempt in this paradigm, ImmerseGAN~\cite{immerseGAN}, uses a GAN-based inpainting architecture~\cite{comodgan} for this task. To achieve text-guided outpainting, ImmerseGAN adopts a pre-trained discriminative network to produce a latent vector representing the given textual description. This latent vector subsequently guides the generator to produce a 360-degree panorama semantically consistent with the text prompt. More recent approaches have primarily focused on leveraging the power of pre-trained latent diffusion models (LDMs) for their strong image generation priors acquired from training on large-scale datasets.

PanoDiff \cite{panodiff}, the first LDM-based method for text-driven NFoV outpainting, is trained on the SUN360 \cite{sun360} dataset. It initially converts the input NFoV images into partial panoramas with visibility masks, and then employs a ControlNet-based LDM \cite{controlnet} for text-guided panorama completion. To ensure geometric continuity at the borders of the generated panorama, PanoDiff further implements a circular padding scheme during inference. Similarly, Diffusion360 \cite{diffusion360} adopts a ControlNet-based LDM \cite{controlnet} to generate 360-degree panoramas from perspective images and textual descriptions. However, instead of circular padding, Diffusion360 leverages a circular blending strategy during the denoising and VAE decoding stages for improved boundary continuity of the generated 360-degree panorama. Recognizing that a single network (as employed in PanoDiff and Diffusion360) often struggles to simultaneously learn the inherent 360-degree panoramic distortion and perform content completion, PanoDecouple~\cite{panodecouple} introduces a decoupled diffusion model as its core contribution. This framework separates the NFoV outpainting process into distortion guidance and content completion. While this modular design is effective, it increases model complexity by requiring a separately trained Distort-CLIP model. Contrastingly, SpND \cite{spnd} incorporates structural prior information from 360-degree panoramic images processed through a spherical network into its diffusion model to guide the 360-degree panoramic image outpainting process.

Depart from the aforementioned methods \cite{panodecouple,panodiff,diffusion360}, which predominantly process and generate 360-degree panoramas using an equirectangular representation throughout their networks, several recent studies have explored leveraging alternative representations for 360-degree panoramic synthesis. CubeDiff \cite{cubediff}, inspired by multi-view diffusion models \cite{mvdream,MVDiffusion}, generates 360-degree panoramas in cubemap format. This cubemap representation enables CubeDiff to more effectively leverage the diffusion priors learned by the LDM from extensive perspective images during the generation process. CubeDiff fine-tunes a pre-trained LDM on a mixed dataset of Structured3D~\cite{structured3d}, Pano360~\cite{pano360}, Polyhaven~\cite{polyhaven}, and Humus~\cite{humus}, using a single conditional view (NFoV image) and textual embeddings as input. Its central innovation is the inflation of 2D attention layers in the LDM into 3D attention layers, enabling the model to explicitly learn inter-face dependencies. This effectively trades the ERP distortion problem for a complex inter-view consistency challenge. Similarly, $\text{Omni}^{2}$~\cite{omni2} produces six overlapping perspective images, each with a 110$\degree$$\times$110$\degree$ field of view, using a Transformer~\cite{transformer} architecture fine-tuned with LoRA on the SUN360~\cite{sun360} and Structured3D~\cite{structured3d} datasets. These overlapping images are subsequently stitched together to synthesize the final 360-degree panoramic image. To address the computational redundancy and resolution constrains introduced by these overlapping images, DreamCube \cite{dreamcube} designs a multi-plane synchronization approach, enabling seamless and consistent RGB-D cubemap generation without overlaps as its main contribution.

\subsubsection{AR-Based}
\label{subsubsec:ar-based}

AOG-Net \cite{aognet} introduces an autoregressive framework, building upon a pre-trained LDM \cite{latentdiffusion}, to progressively outpaint NFoV images into complete panoramas under textual guidance. This stepwise approach enhances the generation of fine-grained visual content and improves alignment with the input textual descriptions. Its key innovation is a global-local conditioning mechanism that integrates multiple guidance signals at each step to ensure consistency across the generated panorama. AOG-Net is trained on Laval Indoor \cite{lavalindoor} and Laval Outdoor \cite{lavaloutdoor} for indoor and outdoor scenarios, respectively. Following the training dataset settings in AOG-Net, OPa-Ma~\cite{opa-ma} uses a pre-trained LDM with Mamba~\cite{mamba}, a state-space model known for its efficiency in handling long sequences, to iteratively outpaint local regions in each step. It introduces two modules: the Visual-textual Consistency Refiner, which enhances input utilization during generation, and the Global-local Mamba Adapter, which ensures global coherence across the generated panorama. In contrast to methods leveraging pre-trained LDMs (such as AOG-Net and OPa-Ma), the recent PAR~\cite{par} adopts a text-conditioned masked autoregressive model~\cite{nova} for 360-degree panoramic image outpainting. Guided by text descriptions, PAR incorporates its designed dual-space circular padding and a translation consistency loss to improve output quality. For all AR-based methods, this step-by-step generation enforces strong logical coherence but at the cost of slower inference and a susceptibility to error propagation.

\subsection{Analysis of Fundamental Design Choices}
\label{subsec:analysis}

The methods surveyed above make fundamentally different design choices that entail significant trade-offs in performance, efficiency, and quality. A critical examination of these choices is essential for understanding the current research landscape and identifying future directions. Below, we systematically analyze the trade-offs and implications of three core architectural decisions: (1) the choice of panoramic representation, (2) the generation framework, and (3) the strategy for adapting pre-trained models.

\subsubsection{Panoramic Representation -- Simplicity vs.\ Prior Compatibility}
\label{subsubsec:choice_of_representation}
As detailed in~\cref{subsec:representations}, methods typically adopt ERP, CMP, or MPP. This selection reflects a trade-off between representational simplicity and compatibility with the generative priors of large pre-trained text-to-image models.
\begin{itemize}
    \item \textbf{Implications of Choosing ERP:} ERP offers a single continuous image, making it directly compatible with standard 2D architectures such as U-Nets and Transformers.
    However, this choice presents two significant challenges. (1) Standard 2D convolutions are not inherently periodic, so to enforce seamlessness at the 360-degree wrap-around boundary, methods must incorporate special treatments such as circular padding (PanoDiff~\cite{panodiff}), circular blending (Diffusion360~\cite{diffusion360}), or dedicated stitching steps (StitchDiffusion~\cite{stitchdiffusion}). (2) ERP introduces non-uniform texel density (geometric distortions), stretching content near the poles and complicating learning of accurate 360-degree panoramas. Some methods address this with spherical manifold convolution (SMGD~\cite{smgd}) or a distortion guidance branch (PanoDecouple~\cite{panodecouple}), albeit at additional computational cost.
    \item \textbf{Implications of Choosing CMP/MPP:} The decision to use a collection of perspective views (CMP or MPP) aims to leverage the priors of pre-trained models more effectively, as these models were trained predominantly on low-distortion perspective images. Representing the 360-degree panoramic scene in this format, as done in CubeDiff~\cite{cubediff} and $\text{Omni}^{2}$~\cite{omni2}, maximizes compatibility and often yields higher-fidelity details. The trade-off is a shift in complexity: the problem of distortion is replaced by the problem of inter-view consistency. This necessitates specialized mechanisms to ensure seamlessness between views, such as inflating 2D attention layers to model 3D relations (CubeDiff~\cite{cubediff}) or employing complex multi-step stitching and inpainting pipelines (PanoFree~\cite{panofree}), which can increase computational overhead and introduce new types of boundary artifacts.
\end{itemize}

\subsubsection{Generation Framework -- NAR-based vs. AR-based} The underlying mechanism for generating 360-degree panoramic content represents a key architectural choice, centered on a trade-off between generation speed and global coherence.
\begin{itemize}
    \item \textbf{Non-Autoregressive (NAR-based):} Most recent methods are built upon non-autoregressive LDMs~\cite{latentdiffusion}, in which the entire latent representation of the 360-degree panorama is denoised in parallel over a series of steps. The primary advantage of this approach is computational efficiency, enabling relatively fast generation of full panoramas, as demonstrated by Diffusion360~\cite{diffusion360}. The main drawback is the difficulty of ensuring global coherence across the vast 360-degree scene. Because the whole image is generated in a single holistic process, the model may struggle to maintain consistent long-range context, sometimes producing mismatched regions or repetitive textures.
    \item \textbf{Autoregressive (AR-based):} In contrast, an autoregressive framework generates the panorama sequentially, explicitly conditioning each new region on the content generated in previous steps. This step-by-step process naturally enforces both local and global consistency, and methods like AOG-Net~\cite{aognet} and PAR~\cite{par} often exhibit strong logical coherence and fine-grained detail. The trade-off for this improved consistency is a substantial increase in inference time, as the sequential process is inherently slower than parallel generation. Moreover, these methods can suffer from error propagation, where low-quality generation in an early step can negatively impact the quality of all subsequent parts.
\end{itemize}

\subsubsection{Adaptation of Pre-trained Models -- Fine-tuning vs. Training-Free}
Since no large-scale generative model is natively trained on 360-degree panoramic image-text pairs, researchers have to adapt models originally trained on standard perspective image-text datasets. Three distinct strategies have emerged, spanning a spectrum of trade-offs among computational cost, data requirements, and model specialization.
\begin{itemize}
    \item \textbf{Full/Substantial Fine-Tuning:} One strategy is to fine-tune a large portion of the model parameters on a specialized 360-degree panoramic image-text dataset. For instance, Diffusion360~\cite{diffusion360} leverages the DreamBooth~\cite{dreambooth} technique to adapt a pre-trained LDM~\cite{latentdiffusion}. This approach enables the model to learn the unique geometric properties and data distributions of 360-degree panoramas, often leading to strong performance. However, the trade-offs are significant: it is usually computationally expensive and time-consuming, requires a substantial panoramic dataset, and may induce catastrophic forgetting, in which the model loses some of the rich generative knowledge from its original training.
    \item \textbf{Parameter-Efficient Fine-Tuning (PEFT):} A more economical alternative trains only a small set of additional parameters while keeping the backbone of the pre-trained model frozen. LoRA~\cite{lora}, employed by methods such as StitchDiffusion~\cite{stitchdiffusion} and DiffPano~\cite{panfusion}, typifies this class. The chief advantage is a drastic reduction in computational and memory requirements, making adaptation more accessible and largely preserving the original model's powerful priors. The main limitation is that the frozen backbone can restrict the model's ability to learn fundamentally new concepts, such as complex spherical geometry, making it potentially less expressive than a fully fine-tuned model.  
    \item \textbf{Training-Free:} The third strategy completely bypasses fine-tuning. Training-free methods like PanoFree~\cite{panofree} and SphereDiff~\cite{spherediff} use pre-trained text-to-image models directly. They achieve panorama generation through careful orchestration of existing capabilities, such as iterative inpainting, view stitching, and specialized sampling algorithms. This approach requires no 360-degree panoramic training data and avoids costly training, fully leveraging the inherent power of the base models. The trade-offs include typically slower inference due to their multi-step, patch-based processes and a higher risk of global inconsistency, as there is no end-to-end training to enforce panoramic coherence.
\end{itemize}

\begin{figure*}
    \centering
    \includegraphics[width=\textwidth]{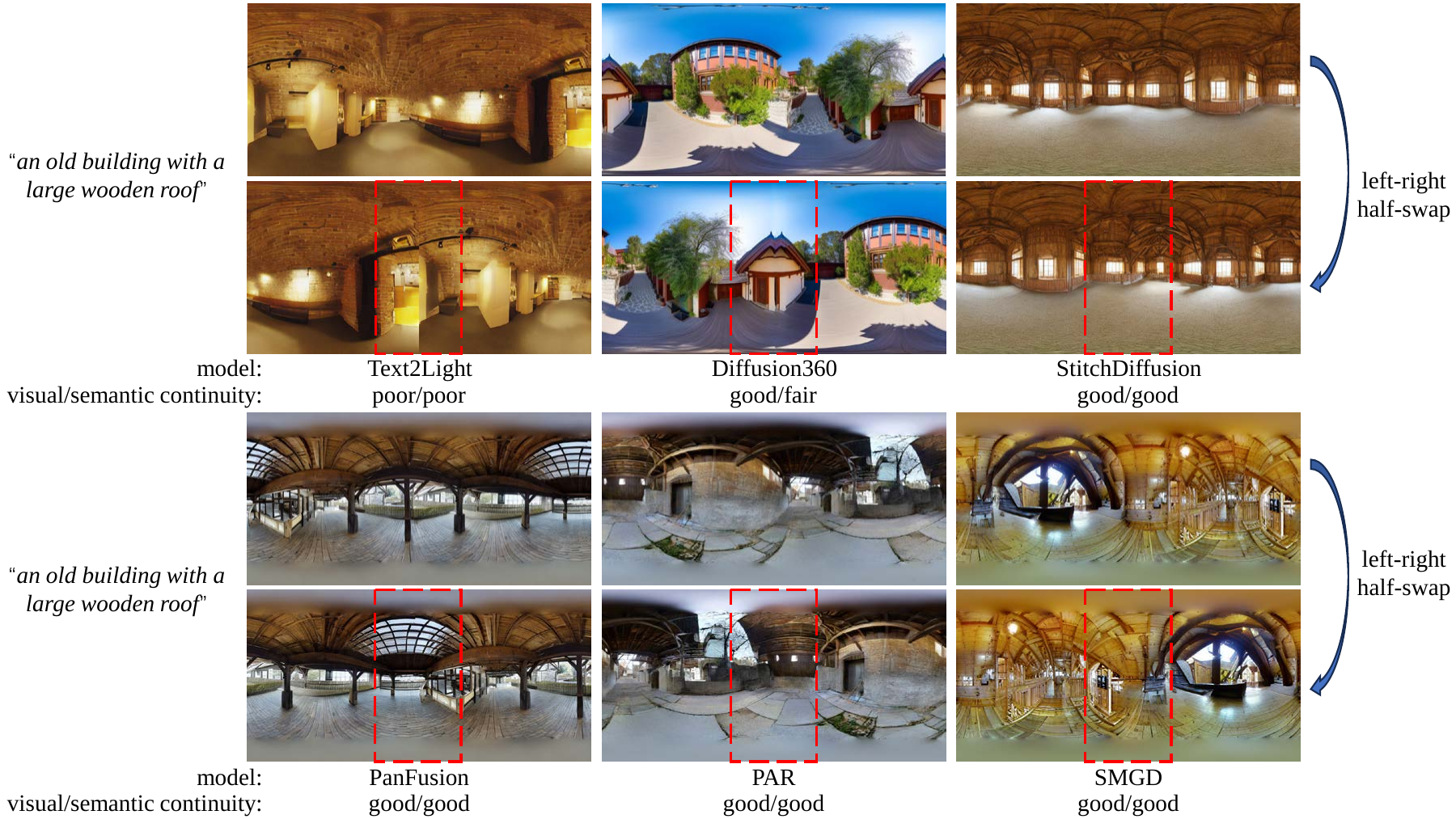}
    \caption{Visual Comparison of Text-Only Generation Methods. To facilitate the assessment of visual and semantic continuities at the image borders,  we generate ``swapped'' versions by exchanging the left and right halves of each generated image, and then use red dashed boxes in these swapped images to highlight the boundary regions. The continuity results are categorized as good, fair, or poor.}
    \label{fig:text_only_comparison}
\end{figure*}

\subsection{Comparisons}

    To systematically evaluate strengths and weaknesses of representative methods, we conduct a benchmark on methods with publicly available inference code and model checkpoints. Our primary objective is to provide a fair and consistent comparison of the official releases, which requires running all models on the same hardware and evaluating them using a consistent set of metrics. We acknowledge that this approach introduces a selection bias, which means our quantitative comparison does not include several recent and important contributions that lack public code or model checkpoints. While this constraint limits the breadth of our comparison, it critically ensures the validity and fairness of the presented results. We encourage readers to consult the original papers for the reported performance of methods not included here. For Text-Only Generation, we compare Text2Light \cite{text2light}, Diffusion360 \cite{diffusion360}, StitchDiffusion \cite{stitchdiffusion}, PanFusion \cite{panfusion}, PAR~\cite{par}, and SMGD~\cite{smgd}. For Text-Driven NFoV Outpainting, we compare PanoDiff \cite{panodiff}, Diffusion360~\cite{diffusion360}, SpND~\cite{spnd} and DreamCube~\cite{dreamcube}.

\begin{figure*}
    \centering
    \includegraphics[width=\textwidth]{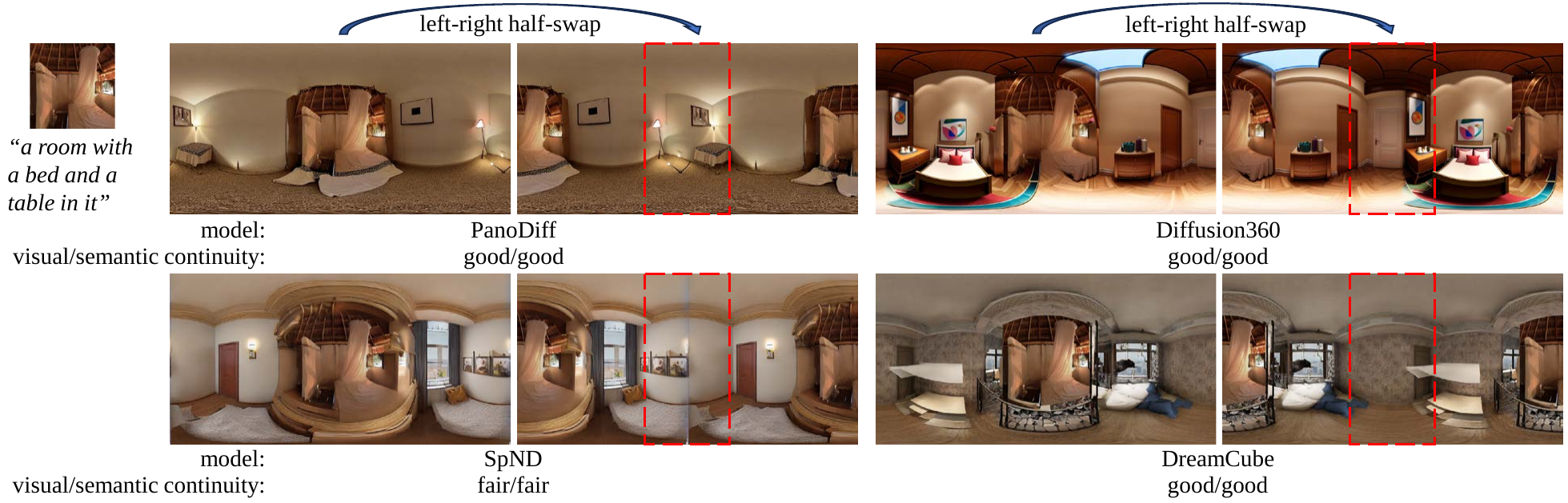}
    \caption{Visual Comparison of Text-Driven NFoV Outpainting Approaches. The rest caption is the same as that for \cref{fig:text_only_comparison}. Note that Diffusion360 provides two separate models. The model for text-driven NFoV outpainting generates images with good semantic continuity at the borders.}
    \label{fig:text_driven_outpainting_comparison}
\end{figure*}

\subsubsection{Qualitative Comparison}
\label{subsec:visual_comparison}

\cref{fig:text_only_comparison} presents a visual comparison of the six text-only generation methods. To facilitate the assessment of visual and semantic continuities at the image borders,  we generate "swapped" versions by exchanging the left and right halves of each generated image. Red dashed boxes are employed in these swapped images to highlight the boundary regions. Among the evaluated methods, Text2Light exhibits obvious seams in the highlighted areas of its swapped images, indicating a failure to maintain visual and semantic continuities at the borders. While Diffusion360 effectively mitigates such observable seams and preserves visual continuity, it occasionally demonstrates local semantic discontinuity at the borders, where content across image boundaries lacks consistent logical or meaningful coherence. In contrast, the other four methods achieve strong visual and semantic continuities at the borders of their synthesized images.

\cref{fig:text_driven_outpainting_comparison} displays the visual comparison for the four text-driven NFoV outpainting approaches. Similar to the text-only comparison, swapped image versions were created to scrutinize the visual and semantic continuities at the borders. PanoDiff, Diffusion360, and DreamCube demonstrate commendable visual and semantic continuities at the borders of the produced images, while SpND exhibits fair visual and semantic continuities.

\begin{table*}
\small
    \centering
    \caption{Quantitative Comparison of Representative Text-Driven 360-Degree Panorama Generation Methods. The first block of rows are for methods in the paradigm of Text-Only Generation, while the second block of rows are for Text-Driven NFoV Outpainting. We use metrics outlined in~\cref{subsec:metrics} for comprehensive evaluation. The inference time and GPU memory, required by each method to generate a 1024$\times$512 360-degree panorama, are also reported. The \textbf{best} and \underline{second-best} results are highlighted for the two paradigms, respectively.}
    \label{tab:comparison}
    { 
\begin{tabular}{lrrrrrrrrr}
\toprule
Method          & FID $\downarrow$ & KID ($\times10^{-2}$) $\downarrow$ & IS $\uparrow$ & CS $\uparrow$  & FAED $\downarrow$ & OmniFID $\downarrow$ & DS $\downarrow$  & Inference (s) $\downarrow$ & GPU Memory (GB) $\downarrow$\\ 
\midrule
Text2Light      & 72.63 & 1.54  & 5.35 & \underline{19.20}    & 18.10       & 99.81  & 5.38 & 33 & 12.5 \\
Diffusion360    & 70.32 & {2.00} & 5.29 & {18.74}    & \textbf{12.43}       & \underline{92.23}  & 0.94 & \underline{3} & \textbf{3.5} \\
StitchDiffusion & 76.69 & 2.04  & \textbf{7.36} & \underline{19.20}    & 15.58       & 108.63  & 1.07 & 28  & \underline{3.6} \\
PanFusion       & \textbf{61.23} & \textbf{1.07}  & 6.16 & 18.96    & \underline{13.16}       & \textbf{92.22}  & 0.85 & 30 & 26.3 \\
PAR       & \underline{64.96} & \underline{1.49}  & \underline{6.68} & 18.91    & 13.99       & 104.02  & \underline{0.76} & 17 & 18.6 \\
SMGD       & 74.91 & 2.00  & 4.23 & \textbf{19.22}    & 16.78       & 106.68  & \textbf{0.75} & \textbf{2} & 8.0 \\
\midrule
PanoDiff & \underline{65.94} & \underline{2.44}  & \underline{4.72} & 19.02    & 10.24       & 122.30  & \underline{1.10} & 48  & 36.0  \\
Diffusion360       & \textbf{64.19} & \textbf{2.05}   & 4.53 & 17.92   & \textbf{5.50}       & \textbf{101.39}  & \textbf{0.72} & \textbf{4} & \textbf{3.7} \\

SpND       & 69.54  & 3.00   & 3.77 & \underline{19.17}   &   \underline{8.67}    & 119.05  & 1.40 & 22 & \underline{16.7} \\

DreamCube       & 66.15 & \textbf{2.05}   & \textbf{4.88} & \textbf{19.26}   & 15.87      & \underline{115.52}  & \underline{1.10} & \underline{12} & \underline{16.7} \\

\bottomrule
\end{tabular}
}
\end{table*}

\subsubsection{Quantitative Comparison}
\label{subsec:comparison}

While the qualitative examples in \cref{fig:text_only_comparison,fig:text_driven_outpainting_comparison} offer valuable visual insights, a quantitative evaluation is essential for a rigorous and objective comparison. To this end, we conduct a comprehensive comparison using metrics outlined in~\cref{subsec:metrics}. 

To ensure an unbiased evaluation of the generalizability of the methods, we employ an out-of-domain dataset, ODI-SR~\cite{lau-net}. This dataset was specifically chosen for two primary reasons: (1) None of the evaluated models were trained on it, which guarantees a fair test of generalization to unseen data; and (2) its diverse composition of indoor and outdoor 360-degree panoramas provides a robust benchmark for evaluating performance across varied real-world scenarios.
For generating text descriptions, we use BLIP2 \cite{blip2} to create textual captions for the 360-degree panoramas included in the ODI-SR dataset. These generated text prompts serve as inputs for both Text-Only Generation and Text-Driven NFoV Outpainting tasks. 
To simulate NFoV images, we first project the equirectangular 360-degree panoramas from the ODI-SR dataset into a cubemap format and then extract the front face of each cubemap. The original 360-degree panoramas from ODI-SR are designated as real images and used as ground truth for the computation of evaluation metrics.

The quantitative results obtained from this comparative evaluation, across the seven evaluation metrics, are presented in \cref{tab:comparison}. To provide insights into the computational efficiency of each method, we also report the inference time and GPU memory required to generate a 1024$\times$512 360-degree panorama on a consistent machine equipped with an RTX A6000 GPU. Note that our results only represent the performance of the publicly released version of each method on a specific GPU (RTX A6000, 48 GB). The actual performance may be influenced by hardware differences and implementation-specific optimizations.
These quantitative findings effectively delineate the strengths and weaknesses of the evaluated approaches, offering valuable guidance for future research in this domain.

\section{Emerging Directions}

This section details two emerging directions that are closely related to text-driven 360-degree panoramic image generation: text-driven 360-degree 3D scene generation and text-driven 360-degree panoramic video generation, respectively.

\subsection{360-Degree 3D Scene Generation} 

Recent advances in text-driven 360-degree panorama generation~\cite{stitchdiffusion,diffusion360,panfusion} have catalyzed innovative methods for reconstructing 360-degree 3D scenes from textual descriptions. 360-degree panoramic images inherently capture both global contexts and geometric constraints of a scene, making them an essential intermediate representation for 3D scene generation. Consequently, recent text-driven 360-degree 3D scene generation methods~\cite{layerpano3d,holodreamer,fastscene,scenedreamer360,dreamscene360,panodreamer,immersegen,perf} use 360-degree panorama generation to bridge the gap between text prompts and 360-degree 3D scene reconstruction. 

\begin{figure*}[t]
    \centering
    \includegraphics[width=\textwidth]{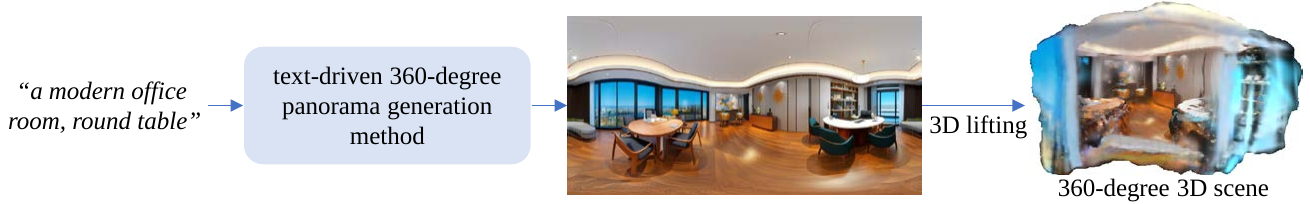}
    \caption{The Framework for Text-Driven 360-Degree 3D Scene Generation using Text-Driven 360-Degree Panorama Generation. This framework accommodates both Text-Only Generation and Text-Driven NFoV Outpainting methods. The input NFoV image is omitted when employing text-driven NFoV outpainting methods for simplicity. ``3D lifting'' denotes the transformation from a generated 360-degree panoramic image to a 3D scene representation by inferring the underlying geometry of the scene.}
    \label{fig:demo2}
\end{figure*}

\begin{table}[t]
\small
\centering
    \caption{Summary of Text-Driven 360-Degree 3D Scene Generation Methods. `360-Degree PG' indicates which text-driven 360-degree panorama generation techniques are adopted to synthesize the intermediate panoramic representation of the scene. `3DGS' denotes 3D Gaussian Splatting.
    }

    \label{tab:t23d}
\resizebox{\linewidth}{!}{
\begin{tabular}{lccc}
\toprule
Method & Publication & 360-Degree PG & 3D Lifting \\ 
\midrule
FastScene      & IJCAI 2024  & Diffusion360    & 3DGS                   \\ 
DreamScene360    & ECCV 2024      & StitchDiffusion         &  3DGS       \\
HoloDreamer         & arxiv 2024        & Diffusion360  & 3DGS \\           
SceneDreamer360    & arxiv 2024    &  PanFusion    &  3DGS      \\
LayerPano3D       & SIGGRAPH 2025        & Diffusion360 \& PanFusion   & 3DGS \\
\bottomrule
\end{tabular}
    }
\end{table}

As depicted in~\cref{fig:demo2}, these methods typically use a two-stage process: (1) 360-Degree Panorama Generation: generating a 360-degree panorama from the input text prompt using a fine-tuned LDM~\cite{latentdiffusion}, and (2) 3D Scene Reconstruction: inferring a 3D representation, typically with 3D Gaussian Splatting (3DGS)~\cite{3dgs}, from the generated panorama and corresponding multi-view perspective images. 
\cref{tab:t23d} provides a comparative summary of the methods using text-driven 360-degree panorama generation techniques introduced in~\cref{sec:methods}.

Within this framework, emerging methods are primarily differentiated by (a) their choice of 360-degree panorama generators and (b) their strategies for extracting and utilizing 3D information.
For instance, FastScene~\cite{fastscene} and HoloDreamer~\cite{holodreamer} both employ Diffusion360~\cite{diffusion360} to generate the initial 360-degree panorama depicting a scene from a given text prompt. FastScene~\cite{fastscene} then synthesizes multi-view panoramas of this scene for specific camera poses using Coarse View Synthesis and Progressive Novel View Inpainting. With these synthesized multi-view panoramas, FastScene introduces Multi-View Projection to get their perspective views. The point clouds derived from these views are then used as input for 3DGS to reconstruct the 3D scene. HoloDreamer~\cite{holodreamer} enhances the Diffusion360-generated panorama with two distinct ControlNet-based LDMs~\cite{controlnet} and a super-resolution network to create a high-resolution, stylized output. Subsequently, HoloDreamer initializes 3D Gaussians using point clouds derived from a reverse equirectangular projection of the high-resolution panorama combined with its corresponding depth information. Finally, a two-stage 3DGS optimization process is developed to refine the scene rendering, resulting in the desired 3D scene reconstruction.

Furthermore, certain methods deviate from the reliance on Diffusion360~\cite{diffusion360}. DreamScene360~\cite{dreamscene360} uses StitchDiffusion~\cite{stitchdiffusion} to generate multiple 360-degree panorama candidates and then employs a self-refinement process to select the optimal candidate for initializing panoramic 3D Gaussians with a 3D geometric field. To facilitate visual feature correspondences between different views and maintain geometric consistencies during the 3DGS optimization process, semantic and geometric regularizations are applied. In contrast, SceneDreamer360~\cite{scenedreamer360} uses a fine-tuned PanFusion~\cite{panfusion} generator, coupled with a super-resolution module from~\cite{diffusion360}, to produce a high-resolution (6K) panorama aligned with the input text prompt. It then uses optimization-based viewpoint selection to extract multi-view images, which are subsequently used for improved point cloud initialization, ultimately leading to 3DGS-based scene reconstruction.

Other methods explore alternative panorama generation techniques. LayerPano3D \cite{layerpano3d} begins by generating four orthogonal perspective views with a fine-tuned text-to-image model \cite{sdxl}. These initial views are then combined with text-guided inpainting \cite{latentdiffusion}, and further processed by using a fine-tuned Diffusion360 \cite{diffusion360} model to outpaint the polar regions, resulting in a reference 360-degree panorama. To handle occlusions in complex scenes, LayerPano3D \cite{layerpano3d} decomposes the reference panorama into multiple depth-based layers and uses a fine-tuned inpainter \cite{panfusion} to complete unseen content at each layer. These inpainted, layered panoramas then provide supervision for panoramic 3D Gaussian scene optimization.

\subsection{360-Degree Panoramic Video Generation}

Analogous to the natural evolution from text-to-image generation ~\cite{imagen,dall-e2,sdxl,latentdiffusion} to text-to-video (T2V) generation~\cite{t2v1,t2v2,t2v3,t2v4}, recent progress in text-driven 360-degree panoramic image synthesis has spurred research into the more challenging task: text-driven 360-degree panoramic video generation. Its representative methods ~\cite{360dvd,dynamicscaler,panodit,videopanda,panowan} are summarized in \cref{tab:t2v}.

\begin{table}[t]
\small
\centering
    \caption{Summary of Text-Driven 360-Degree Panoramic Video Generation Methods. `TF' specifies if it is training-free.
    }

    \label{tab:t2v}
\resizebox{\linewidth}{!}{
\begin{tabular}{lcccc}
\toprule
Method & Publication & Training Dataset & TF & Representation \\ 
\midrule
360DVD~\cite{360dvd}     & CVPR 2024  & WEB360~\cite{360dvd}      & $\times$ & ERP                   \\
DynamicScaler~\cite{dynamicscaler}         & CVPR 2025        & N/A  & $\checkmark$  & ERP \\ 
PanoDiT~\cite{panodit}   & AAAI 2025    &  PHQ360~\cite{panodit}    &  $\times$  & ERP    \\
SphereDiff~\cite{spherediff}   & arxiv 2025    &  N/A    &  $\checkmark$  & Spherical    \\
VideoPanda~\cite{videopanda}   & arxiv 2025    &  WEB360~\cite{360dvd}    &  $\times$  & MPP    \\
PanoWan~\cite{panowan}   & arxiv 2025    &  PanoVid~\cite{panowan}    &  $\times$  & ERP    \\
\bottomrule
\end{tabular}
    }
\end{table}

Most works in this area rely on specialized training. 360DVD~\cite{360dvd} pioneers this task by first constructing a tailored dataset, WEB360, consisting of 2,114 360-degree panoramic video-text pairs. Using this dataset, it trains an adapter to enable a pre-trained T2V models \cite{t2v2} to synthesize 360-degree panoramic videos from provided text prompts. Inspired by PanoDiff~\cite{panodiff}, 360DVD adopts a latent rotation mechanism in the inference process to maintain the boundary continuity of the synthesized results. Limitations in 360DVD, particularly the lack of detailed motion descriptions in its dataset, have prompted further refinements. PanoDiT~\cite{panodit} addresses this by curating a higher-quality subset named PHQ360 and replacing the U-Net architecture with a Diffusion Transformer (DiT)~\cite{dit} and a motion LoRA for improved generation. Similarly, the rotation mechanism from PanoDiff is used in the post-processing phase of PanoDiT to ensure continuity. Recognizing the critical role of large-scale and high-quality data, PanoWan~\cite{panowan} establishes PanoVid, a dataset with over 13,000 video-text pairs. It introduces a latitude-aware sampling technique to mitigate ERP distortions and fined-tune a DiT-based T2V model~\cite{wan} with LoRA for generation.

In a different vein, VideoPanda~\cite{videopanda} seeks to better leverage priors from pre-trained T2V models. Trained on WEB360, it introduces multi-view attention layers to synthesize multiple perspective video outputs, which are then stitched together to form the final 360-degree panoramic video. This approach avoids direction in the equirectangular projection space.

Contrasting with these trained-based methods, several approaches have explored trained-free generation. DynamicScaler~\cite{dynamicscaler} designs an offset-shifting denoiser and a panoramic projection technique to synthesize a low-resolution 360-degree panoramic video, which then provides global motion guidance for refining a high-resolution version. However, its reliance on the ERP latent representation can lead to discontinuities near the poles. To address this, SphereDiff~\cite{spherediff} introduces a spherical latent representation and extends Multidiffusion~\cite{multidiffusion} to the constructed spherical space, achieving a more uniform distribution and improving quality at the poles. A common challenge to these training-free methods is that their patch-based synthesis mechanism can introduce global inconsistencies. Future work could focus on incorporating global guidance into these frameworks to mitigate this issue.

\condensedpara{Datasets for 360-Degree Panoramic Video Generation.} 
While the datasets discussed in~\cref{subsec:datasets} are foundational for static 360-degree panoramic image generation, the task of 360-degree panoramic video generation requires specialized datasets that include temporal information. These datasets are crucial for training and evaluating models capable of producing coherent and immersive 360-degree panoramic video content. Key datasets in this domain include:
\begin{itemize}
    \item \textbf{WEB360}~\cite{360dvd} offers 2,114 text-video pairs of 360-degree panoramas. The videos are sourced from existing datasets such as ODV360~\cite{odv360} and platforms like YouTube. To generate detailed textual descriptions for the videos, a combination of BLIP~\cite{blip} and ChatGPT was employed.  
    \item \textbf{YouTube360}~\cite{imagine360} provides 9,557 360-degree panoramic videos sourced from YouTube, featuring diverse scenes such as virtual city tours and wildlife documentaries. The corresponding text prompts were generated using VideoLLaMa-2~\cite{videollama2}.
    \item \textbf{360-1M}~\cite{360-1m} is a large-scale dataset consisting of 1,076,592 360-degree videos, collected from YouTube and distributed across 15 categories. As it was not originally created for text-driven generation, this dataset does not provide paired textual descriptions.
    \item \textbf{PanoVid}~\cite{panowan} is a high-quality dataset of over 13,000 video clips curated specifically for text-driven 360-degree panoramic video generation. The videos in PanoVid are collected from multiple sources, including WEB360, YouTube360, and 360-1M. Qwen-2.5-VL~\cite{qwen} was adopted to produce rich textual descriptions for the video content.
\end{itemize}

\section{Challenges and Future Directions}

Despite the impressive results achieved in text-driven 360-degree panorama generation, challenges remain in evaluation metrics, resolution, controllability, model design, societal impact and industrial adoption. This section identifies these challenges and outlines potential directions for future research.

\paragraph{Evaluation Metrics} \label{sec:future_metrics}
As established in our analysis (see~\cref{subsec:metrics}), the development of metrics for global scene consistency remains a key challenge. This includes creating more robust, panorama-aware methods for evaluating both text-to-image semantic alignment and the overall structural plausibility of the generated 360-degree panoramic space. Future work could explore using advanced Vision-Language Models (VLMs) for question-based evaluations of complex spatial relationships. Another promising direction is the development of metrics that assess the implicit 3D geometry of the scene to detect logical inconsistencies in scale and layout that 2D metrics currently miss.

\paragraph{Higher Resolution} While Diffusion360~\cite{diffusion360}, which uses a super-resolution module, is among the few methods that can currently achieve a maximum resolution of 6144$\times$3072 (6K), this remains inadequate for demanding applications like VR gaming and high-fidelity 3D scene reconstruction, which often necessitate resolutions of 8K or higher to capture intricate details of landscapes and architecture. However, generating such high-resolution panoramas incurs both high memory consumption and long inference times, which severely limit practical deployment. Addressing these limitations will require the development of more efficient model architectures and optimization techniques. Promising approaches include the use of window-based operations, model pruning, quantization, knowledge distillation, and advanced neural network designs tailored to resource-intensive tasks. Moreover, the availability of high-resolution, large-scale datasets will be critical for driving progress in this direction.

\paragraph{Multi-modal Generation} Existing text-driven methods, despite their ability to produce photorealistic 360-degree panoramas, often lack precise control over global semantic layout and spatial structure of the generated scene. This motivates exploring multi-modal approaches to enhance controllability. Although 360PanT \cite{360pant} demonstrates panorama-to-panorama translation using auxiliary modalities like edge and segmentation maps alongside text, its outputs deviate from the standard 360$\degree$$\times$180$\degree$ field of view when these additional modalities are incorporated. Future research should focus on developing multi-modal techniques that effectively integrate diverse inputs (\textit{e.g.}~depth maps, segmentation maps, or edge maps) with text prompts to achieve fine-grained spatial control in the generated 360-degree panoramas, while ensuring strict adherence to the standard equirectangular projection format.

\paragraph{Model Design} Most text-driven 360-degree panorama generation methods are built upon latent diffusion models (LDMs) \cite{latentdiffusion}. While LDMs have achieved remarkable success in text-to-image synthesis, recent advancements in autoregressive models indicate promising alternative architectures. Specifically, Visual Autoregressive (VAR) models, exemplified by Infinity~\cite{infinity}, have exhibited superior performance compared to the leading LDMs in standard text-to-image synthesis. This highlights an exciting avenue for future research: exploring VAR-based models for text-driven 360-degree panorama generation.

\paragraph{Ethical and Societal Considerations} The persuasive nature of 360-degree panoramas introduces risks of misuse, including fabricated environments, disinformation, and privacy violations. To mitigate these concerns, future research should pair technical advances with safeguards such as transparent data documentation, responsible licensing, and watermarking mechanisms. Proactive engagement with ethical guidelines and interdisciplinary oversight is essential to ensure that text-driven 360-degree panorama generation benefits society while limiting potential harm.

\paragraph{Industry Translation and Adoption}
    Text-driven 360-degree panorama generation is beginning to move from research to practice, supporting VR/AR content creation, virtual tourism, and game development. Major industry players, including Meta (Quest), Google (Street View), and Apple (Vision Pro), are actively developing related capabilities. Broader deployment is currently limited by requirements for stable 8K plus resolution, low latency on-device inference, and industry-standard benchmarks. Addressing these challenges through more efficient model design and closer collaboration between academia and industry will be critical for production-ready adoption.

\section{Conclusion}

This survey has provided a comprehensive overview of the rapidly evolving field of text-driven 360-degree panoramic image generation. We began by introducing primary representation methods of 360-degree panoramic images, along with widely used datasets, and key evaluation metrics in this domain. Subsequently, we presented an in-depth discussion of prevalent methods for text-driven 360-degree panorama generation, and explored its two closely related directions: text-driven 360-degree 3D scene generation and text-driven 360-degree panoramic video synthesis. Despite the significant progress achieved in this field, several challenges remain. To address these challenges, we have articulated promising directions for future research.


\begin{thebibliography}{1}
\bibliographystyle{IEEEtran}

\bibitem{lau-net}
X.~Deng, H.~Wang, M.~Xu, Y.~Guo, Y.~Song, and L.~Yang, ``Lau-net: Latitude
  adaptive upscaling network for omnidirectional image super-resolution,'' in
  \emph{Proceedings of the IEEE/CVF Conference on Computer Vision and Pattern
  Recognition}, 2021, pp. 9189--9198.

\bibitem{infinity}
J.~Han, J.~Liu, Y.~Jiang, B.~Yan, Y.~Zhang, Z.~Yuan, B.~Peng, and X.~Liu,
  ``Infinity: Scaling bitwise autoregressive modeling for high-resolution image
  synthesis,'' in \emph{Proceedings of the IEEE/CVF Conference on Computer
  Vision and Pattern Recognition}, 2025, pp.~15733--15744.

\bibitem{mamba}
A.~Gu and T.~Dao, ``Mamba: Linear-time sequence modeling with selective state
  spaces,'' in \emph{First Conference on Language Modeling}, 2024.

\bibitem{opa-ma}
P.~Gao, K.~Yao, T.~Ye, S.~Wang, Y.~Yao, and X.~Wang, ``Opa-ma: Text guided
  mamba for 360-degree image out-painting,'' \emph{arXiv preprint
  arXiv:2407.10923}, 2024.

\bibitem{polyhaven}
{Poly Haven}, ``{HDRIs},'' \url{https://polyhaven.com/hdris}, 2025, accessed:
  February, 2025.

\bibitem{humus}
E.~Persson, ``Texture from humus,''
  \url{https://www.humus.name/index.php?page=Textures}, 2025, accessed:
  February, 2025.

\bibitem{360survey}
T.~L. da~Silveira, P.~G. Pinto, J.~Murrugarra-Llerena, and C.~R. Jung, ``3d
  scene geometry estimation from 360 imagery: A survey,'' \emph{ACM Computing
  Surveys}, vol.~55, no.~4, pp. 1--39, 2022.

\bibitem{nerf}
B.~Mildenhall, P.~P. Srinivasan, M.~Tancik, J.~T. Barron, R.~Ramamoorthi, and
  R.~Ng, ``Nerf: Representing scenes as neural radiance fields for view
  synthesis,'' in \emph{European Conference on Computer Vision}.\hskip 1em plus
  0.5em minus 0.4em\relax Springer, 2020, pp. 405--421.

\bibitem{sdf}
J.~J. Park, P.~Florence, J.~Straub, R.~Newcombe, and S.~Lovegrove, ``Deepsdf:
  Learning continuous signed distance functions for shape representation,'' in
  \emph{Proceedings of the IEEE/CVF conference on computer vision and pattern
  recognition}, 2019, pp. 165--174.

\bibitem{occupancy}
L.~Mescheder, M.~Oechsle, M.~Niemeyer, S.~Nowozin, and A.~Geiger, ``Occupancy
  networks: Learning 3d reconstruction in function space,'' in
  \emph{Proceedings of the IEEE/CVF conference on computer vision and pattern
  recognition}, 2019, pp. 4460--4470.

\bibitem{glide}
A.~Q. Nichol, P.~Dhariwal, A.~Ramesh, P.~Shyam, P.~Mishkin, B.~Mcgrew,
  I.~Sutskever, and M.~Chen, ``Glide: Towards photorealistic image generation
  and editing with text-guided diffusion models,'' in \emph{International
  Conference on Machine Learning}.\hskip 1em plus 0.5em minus 0.4em\relax PMLR,
  2022, pp. 16\,784--16\,804.

\bibitem{imagen}
C.~Saharia, W.~Chan, S.~Saxena, L.~Li, J.~Whang, E.~L. Denton, K.~Ghasemipour,
  R.~Gontijo~Lopes, B.~Karagol~Ayan, T.~Salimans \emph{et~al.},
  ``Photorealistic text-to-image diffusion models with deep language
  understanding,'' \emph{Advances in neural information processing systems},
  vol.~35, pp. 36\,479--36\,494, 2022.

\bibitem{dall-e2}
A.~Ramesh, P.~Dhariwal, A.~Nichol, C.~Chu, and M.~Chen, ``Hierarchical
  text-conditional image generation with clip latents,'' \emph{arXiv preprint
  arXiv:2204.06125}, vol.~1, no.~2, p.~3, 2022.

\bibitem{immerseGAN}
M.~R.~K. Dastjerdi, Y.~Hold-Geoffroy, J.~Eisenmann, S.~Khodadadeh, and J.-F.
  Lalonde, ``Guided co-modulated gan for 360 field of view extrapolation,'' in
  \emph{2022 International Conference on 3D Vision (3DV)}.\hskip 1em plus 0.5em
  minus 0.4em\relax IEEE, 2022, pp. 475--485.

  
\bibitem{HM3D}
S.~K. Ramakrishnan, A.~Gokaslan, E.~Wijmans, O.~Maksymets, A.~Clegg, J.~M.
  Turner, E.~Undersander, W.~Galuba, A.~Westbury, A.~X. Chang \emph{et~al.},
  ``Habitat-matterport 3d dataset (hm3d): 1000 large-scale 3d environments for
  embodied ai,'' in \emph{Thirty-fifth Conference on Neural Information
  Processing Systems Datasets and Benchmarks Track (Round 2)}, 2021.

\bibitem{llama2}
H.~Touvron, L.~Martin, K.~Stone, P.~Albert, A.~Almahairi, Y.~Babaei,
  N.~Bashlykov, S.~Batra, P.~Bhargava, S.~Bhosale \emph{et~al.}, ``Llama 2:
  Open foundation and fine-tuned chat models,'' \emph{arXiv preprint
  arXiv:2307.09288}, 2023.

\bibitem{blip}
J.~Li, D.~Li, C.~Xiong, and S.~Hoi, ``Blip: Bootstrapping language-image
  pre-training for unified vision-language understanding and generation,'' in
  \emph{International conference on machine learning}.\hskip 1em plus 0.5em
  minus 0.4em\relax PMLR, 2022, pp. 12\,888--12\,900.

\bibitem{blip2}
J.~Li, D.~Li, S.~Savarese, and S.~Hoi, ``Blip-2: Bootstrapping language-image
  pre-training with frozen image encoders and large language models,'' in
  \emph{International conference on machine learning}.\hskip 1em plus 0.5em
  minus 0.4em\relax PMLR, 2023, pp. 19\,730--19\,742.

\bibitem{pano360}
M.~Kocabas, C.-H.~P. Huang, J.~Tesch, L.~M{\"u}ller, O.~Hilliges, and M.~J.
  Black, ``Spec: Seeing people in the wild with an estimated camera,'' in
  \emph{Proceedings of the IEEE/CVF International Conference on Computer
  Vision}, 2021, pp. 11\,035--11\,045.

\bibitem{kid}
M.~Bi{\'n}kowski, D.~J. Sutherland, M.~Arbel, and A.~Gretton, ``Demystifying
  mmd gans,'' in \emph{International Conference on Learning Representations},
  2018.

\bibitem{sdxl}
D.~Podell, Z.~English, K.~Lacey, A.~Blattmann, T.~Dockhorn, J.~Müller, J.~Penna,
  and R.~Rombach, ``SDXL: Improving latent diffusion models for high-resolution
  image synthesis,'' in \emph{Proceedings of the International Conference on
  Learning Representations}, 2024.


\bibitem{spherediff}
M.~Park, T.~Kang, J.~Yun, S.~Hwang, and J.~Choo, “SphereDiff: Tuning-free Omnidirectional Panoramic Image and Video Generation via Spherical Latent Representation,” \emph{arXiv preprint arXiv:2504.14396}, 2025.

\bibitem{omni2}
Y.~Yang, H.~Duan, Y.~Zhu, X.~Liu, L.~Liu, Z.~Xu, G.~Ma, X.~Min, G.~Zhai, and P.~L.~Callet, “Omni\textsuperscript{2}: Unifying Omnidirectional Image Generation and Editing in an Omni Model,” \emph{arXiv preprint arXiv:2504.11379}, 2025.


\bibitem{panodecouple}
D.~Zheng, C.~Zhang, X.-M. Wu, C.~Li, C.~Lv, J.-F. Hu, and W.-S. Zheng,
  ``Panorama generation from NFoV image done right,'' in \emph{Proceedings of
  the IEEE/CVF Conference on Computer Vision and Pattern Recognition}, 2025,
  pp.~21610--21619.



\bibitem{par}
C.~Wang, X.~Li, L.~Qi, X.~Lin, J.~Bai, Q.~Zhou, and Y.~Tong, “Conditional Panoramic Image Generation via Masked Autoregressive Modeling,” \emph{arXiv preprint arXiv:2505.16862}, 2025.


\bibitem{3dgs}
B.~Kerbl, G.~Kopanas, T.~Leimk{\"u}hler, and G.~Drettakis, ``3d gaussian
  splatting for real-time radiance field rendering.'' \emph{ACM Trans. Graph.},
  vol.~42, no.~4, pp. 139--1, 2023.

\bibitem{layerpano3d}
S.~Yang, J.~Tan, M.~Zhang, T.~Wu, G.~Wetzstein, Z.~Liu, and D.~Lin,
  ``LayerPano3D: Layered 3D panorama for hyper-immersive scene generation,''
  in \emph{Proceedings of the ACM SIGGRAPH Conference Papers}, 2025,
  pp.~1--10.


\bibitem{360pant}
H.~Wang and J.-H. Xue, ``360pant: Training-free text-driven 360-degree
  panorama-to-panorama translation,'' in \emph{2025 IEEE/CVF Winter Conference
  on Applications of Computer Vision (WACV)}.\hskip 1em plus 0.5em minus
  0.4em\relax IEEE, 2025, pp. 212--221.

\bibitem{3dgen}
W.~Xia and J.-H. Xue, ``A survey on deep generative 3d-aware image synthesis,''
  \emph{ACM Computing Surveys}, vol.~56, no.~4, pp. 1--34, 2023.

\bibitem{holodreamer}
H.~Zhou, X.~Cheng, W.~Yu, Y.~Tian, and L.~Yuan, ``Holodreamer: Holistic 3d
  panoramic world generation from text descriptions,'' \emph{arXiv preprint
  arXiv:2407.15187}, 2024.


\bibitem{fastscene}
Y.~Ma, D.~Zhan, and Z.~Jin, ``FastScene: Text-driven fast 3D indoor scene
  generation via panoramic Gaussian splatting,'' in \emph{Proceedings of the
  International Joint Conference on Artificial Intelligence}, 2024,
  pp.~1173--1181.



\bibitem{scenedreamer360}
W.~Li, F.~Cai, Y.~Mi, Z.~Yang, W.~Zuo, X.~Wang, and X.~Fan, ``Scenedreamer360:
  Text-driven 3d-consistent scene generation with panoramic gaussian
  splatting,'' \emph{arXiv preprint arXiv:2408.13711}, 2024.

\bibitem{dreamscene360}
S.~Zhou, Z.~Fan, D.~Xu, H.~Chang, P.~Chari, T.~Bharadwaj, S.~You, Z.~Wang, and
  A.~Kadambi, ``Dreamscene360: Unconstrained text-to-3d scene generation with
  panoramic gaussian splatting,'' in \emph{European Conference on Computer
  Vision}.\hskip 1em plus 0.5em minus 0.4em\relax Springer, 2024, pp. 324--342.

\bibitem{autoencoder}
G.~E. Hinton and R.~R. Salakhutdinov, ``Reducing the dimensionality of data
  with neural networks,'' \emph{science}, vol. 313, no. 5786, pp. 504--507,
  2006.

\bibitem{tcsvtt2i2}
H.~Chen, Y.~Zhang, X.~Wang, X.~Duan, Y.~Zhou, and W.~Zhu, ``DisenDreamer: Subject-driven text-to-image generation with sample-aware disentangled tuning,'' \emph{IEEE Transactions on Circuits and Systems for Video Technology}, 2024.


\bibitem{inception}
C.~Szegedy, V.~Vanhoucke, S.~Ioffe, J.~Shlens, and Z.~Wojna, ``Rethinking the
  inception architecture for computer vision,'' in \emph{Proceedings of the
  IEEE conference on computer vision and pattern recognition}, 2016, pp.
  2818--2826.

\bibitem{controlnet}
L.~Zhang, A.~Rao, and M.~Agrawala, ``Adding conditional control to
  text-to-image diffusion models,'' in \emph{Proceedings of the IEEE/CVF
  international conference on computer vision}, 2023, pp. 3836--3847.

\bibitem{dreambooth}
N.~Ruiz, Y.~Li, V.~Jampani, Y.~Pritch, M.~Rubinstein, and K.~Aberman,
  ``Dreambooth: Fine tuning text-to-image diffusion models for subject-driven
  generation,'' in \emph{Proceedings of the IEEE/CVF conference on computer
  vision and pattern recognition}, 2023, pp. 22\,500--22\,510.

\bibitem{lora}
E.~J. Hu, Y.~Shen, P.~Wallis, Z.~Allen-Zhu, Y.~Li, S.~Wang, L.~Wang, W.~Chen
  \emph{et~al.}, ``Lora: Low-rank adaptation of large language models.''
  \emph{ICLR}, vol.~1, no.~2, p.~3, 2022.


\bibitem{diffpano}
W.~Ye, C.~Ji, Z.~Chen, J.~Gao, X.~Huang, S.-H. Zhang, W.~Ouyang, T.~He,
  C.~Zhao, and G.~Zhang, ``DiffPano: Scalable and consistent text to panorama
  generation with spherical epipolar-aware diffusion,'' \emph{Advances in
  Neural Information Processing Systems}, vol.~37, pp.~1304--1332, 2024.


\bibitem{panofree}
A.~Liu, Z.~Li, Z.~Chen, N.~Li, Y.~Xu, and B.~A. Plummer, ``Panofree:
  Tuning-free holistic multi-view image generation with cross-view
  self-guidance,'' in \emph{European Conference on Computer Vision}.\hskip 1em
  plus 0.5em minus 0.4em\relax Springer, 2024, pp. 146--164.

\bibitem{panodiff}
J.~Wang, Z.~Chen, J.~Ling, R.~Xie, and L.~Song, ``360-degree panorama
  generation from few unregistered nfov images,'' in \emph{Proceedings of the
  31st ACM International Conference on Multimedia}, 2023, pp. 6811--6821.

\bibitem{vqgan}
P.~Esser, R.~Rombach, and B.~Ommer, ``Taming transformers for high-resolution
  image synthesis,'' in \emph{Proceedings of the IEEE/CVF conference on
  computer vision and pattern recognition}, 2021, pp. 12\,873--12\,883.

\bibitem{cubediff}
N.~Kalischek, M.~Oechsle, F.~Manhardt, P.~Henzler, K.~Schindler, and
  F.~Tombari, ``Cubediff: Repurposing diffusion-based image models for panorama
  generation,'' in \emph{The Thirteenth International Conference on Learning
  Representations}, 2025.

\bibitem{lavalindoor}
M.-A. Gardner, K.~Sunkavalli, E.~Yumer, X.~Shen, E.~Gambaretto, C.~Gagn{\'e},
  and J.-F. Lalonde, ``Learning to predict indoor illumination from a single
  image,'' \emph{ACM Transactions on Graphics (TOG)}, vol.~36, no.~6, pp.
  1--14, 2017.

\bibitem{lavaloutdoor}
Y.~Hold-Geoffroy, A.~Athawale, and J.-F. Lalonde, ``Deep sky modeling for
  single image outdoor lighting estimation,'' in \emph{Proceedings of the
  IEEE/CVF conference on computer vision and pattern recognition}, 2019, pp.
  6927--6935.


\bibitem{structured3d}
J.~Zheng, J.~Zhang, J.~Li, R.~Tang, S.~Gao, and Z.~Zhou,
  ``Structured3D: A large photo-realistic dataset for structured 3D modeling,''
  in \emph{Proceedings of the European Conference on Computer Vision}, 2020,
  pp.~519--535.



\bibitem{sun360}
J.~Xiao, K.~A. Ehinger, A.~Oliva, and A.~Torralba, ``Recognizing scene
  viewpoint using panoramic place representation,'' in \emph{2012 IEEE
  conference on computer vision and pattern recognition}.\hskip 1em plus 0.5em
  minus 0.4em\relax IEEE, 2012, pp. 2695--2702.


\bibitem{matterport3d}
A.~Chang, A.~Dai, T.~Funkhouser, M.~Halber, M.~Niebner, M.~Savva, S.~Song,
  A.~Zeng, and Y.~Zhang, ``Matterport3D: Learning from RGB-D data in indoor
  environments,'' in \emph{Proceedings of the International Conference on 3D
  Vision (3DV)}, 2017, pp.~667--676.


\bibitem{faed}
C.~Oh, W.~Cho, Y.~Chae, D.~Park, L.~Wang, and K.-J. Yoon, ``Bips: Bi-modal
  indoor panorama synthesis via residual depth-aided adversarial learning,'' in
  \emph{European Conference on Computer Vision}.\hskip 1em plus 0.5em minus
  0.4em\relax Springer, 2022, pp. 352--371.

\bibitem{omnifid}
A.~Christensen, N.~Mojab, K.~Patel, K.~Ahuja, Z.~Akata, O.~Winther,
  M.~Gonzalez-Franco, and A.~Colaco, ``Geometry fidelity for spherical
  images,'' in \emph{European Conference on Computer Vision}.\hskip 1em plus
  0.5em minus 0.4em\relax Springer, 2024, pp. 276--292.

\bibitem{clip}
A.~Radford, J.~W. Kim, C.~Hallacy, A.~Ramesh, G.~Goh, S.~Agarwal, G.~Sastry,
  A.~Askell, P.~Mishkin, J.~Clark \emph{et~al.}, ``Learning transferable visual
  models from natural language supervision,'' in \emph{International conference
  on machine learning}.\hskip 1em plus 0.5em minus 0.4em\relax PmLR, 2021, pp.
  8748--8763.


\bibitem{fid}
M.~Heusel, H.~Ramsauer, T.~Unterthiner, B.~Nessler, and S.~Hochreiter, 
``GANs trained by a two time-scale update rule converge to a local Nash equilibrium,'' 
in \emph{Advances in Neural Information Processing Systems}, 
vol.~30, pp.~6626--6637, 2017.


\bibitem{is}
T.~Salimans, I.~Goodfellow, W.~Zaremba, V.~Cheung, A.~Radford, and X.~Chen, 
``Improved techniques for training GANs,'' 
in \emph{Advances in Neural Information Processing Systems}, 
vol.~29, pp.~2234--2242, 2016.


\bibitem{multidiffusion}
O.~Bar-Tal, L.~Yariv, Y.~Lipman, and T.~Dekel, ``Multidiffusion: Fusing
  diffusion paths for controlled image generation,'' in \emph{International
  Conference on Machine Learning}.\hskip 1em plus 0.5em minus 0.4em\relax PMLR,
  2023, pp. 1737--1752.

\bibitem{text2light}
Z.~Chen, G.~Wang, and Z.~Liu, ``Text2light: Zero-shot text-driven hdr panorama
  generation,'' \emph{ACM Transactions on Graphics (TOG)}, vol.~41, no.~6, pp.
  1--16, 2022.

\bibitem{stitchdiffusion}
H.~Wang, X.~Xiang, Y.~Fan, and J.-H. Xue, ``Customizing 360-degree panoramas
  through text-to-image diffusion models,'' in \emph{Proceedings of the
  IEEE/CVF Winter Conference on Applications of Computer Vision}, 2024, pp.
  4933--4943.

\bibitem{diffusion360}
M.~Feng, J.~Liu, M.~Cui, and X.~Xie, ``Diffusion360: Seamless 360 degree
  panoramic image generation based on diffusion models,'' \emph{arXiv preprint
  arXiv:2311.13141}, 2023.

\bibitem{panfusion}
C.~Zhang, Q.~Wu, C.~C. Gambardella, X.~Huang, D.~Phung, W.~Ouyang, and J.~Cai,
  ``Taming stable diffusion for text to 360 panorama image generation,'' in
  \emph{Proceedings of the IEEE/CVF Conference on Computer Vision and Pattern
  Recognition}, 2024, pp. 6347--6357.

\bibitem{aognet}
Z.~Lu, K.~Hu, C.~Wang, L.~Bai, and Z.~Wang, ``Autoregressive omni-aware
  outpainting for open-vocabulary 360-degree image generation,'' in
  \emph{Proceedings of the AAAI Conference on Artificial Intelligence},
  vol.~38, no.~13, 2024, pp. 14\,211--14\,219.

\bibitem{MVDiffusion}
S.~Tang, F.~Zhang, J.~Chen, P.~Wang, and Y.~Furukawa, ``Mvdiffusion: Enabling
  holistic multi-view image generation with correspondence-aware diffusion,''
  in \emph{Thirty-seventh Conference on Neural Information Processing Systems},
  2023.

\bibitem{mvdream}
Y.~Shi, P.~Wang, J.~Ye, L.~Mai, K.~Li, and X.~Yang, ``Mvdream: Multi-view
  diffusion for 3d generation,'' in \emph{The Twelfth International Conference
  on Learning Representations}, 2024.

\bibitem{latentdiffusion}
R.~Rombach, A.~Blattmann, D.~Lorenz, P.~Esser, and B.~Ommer, ``High-resolution
  image synthesis with latent diffusion models,'' in \emph{Proceedings of the
  IEEE/CVF conference on computer vision and pattern recognition}, 2022, pp.
  10\,684--10\,695.

\bibitem{tcsvtt2i}
Z.~Wang, O.~Li, T.~Wang, L.~Wei, Y.~Hao, X.~Wang, and Q.~Tian, ``Prior Preserved Text-to-Image Personalization without Image Regularization,'' \emph{IEEE Transactions on Circuits and Systems for Video Technology}, 2024.


\bibitem{t2i-survey}
C.~Zhang, C.~Zhang, M.~Zhang, and I.~S.~Kweon, “Text-to-image diffusion models in generative AI: A survey,” \emph{arXiv preprint arXiv:2303.07909}, 2023.

\bibitem{t2i-survey2}
F.-A. Croitoru, V.~Hondru, R.~T. Ionescu, and M.~Shah, ``Diffusion models in
vision: A survey,'' \emph{IEEE Transactions on Pattern Analysis and Machine
Intelligence}, vol.~45, no.~9, pp.~10850--10869, 2023.

\bibitem{t2i-survey3}
F.~Bie, Y.~Yang, Z.~Zhou, A.~Ghanem, M.~Zhang, Z.~Yao, X.~Wu, C.~Holmes,
  P.~Golnari, D.~A. Clifton \emph{et al.}, ``RenAIssance: A survey into AI
  text-to-image generation in the era of large model,'' \emph{IEEE Transactions
  on Pattern Analysis and Machine Intelligence}, vol.~47, no.~3, pp.~2212--2231,
  2025.


\bibitem{hdri-skies}
{HDRI Skies}, ``Free HDRI Skies,'' \url{https://hdri-skies.com/}, 2025, accessed:
 February, 2025.

\bibitem{hdrmaps}
{HDRMAPS}, ``HDRI Maps and Textures,'' \url{https://hdrmaps.com/}, 2025, accessed:
 February, 2025.

\bibitem{ihdri}
{iHDRI}, ``HDRI Skies – Outdoor,'' \url{https://www.ihdri.com/hdri-skies-outdoor/}, accessed:
 February, 2025.

\bibitem{360cities}
{360Cities}, ``360Cities – World Panoramic Photography,'' \url{https://www.360cities.net/}, accessed:
 February, 2025.

\bibitem{tcsvtsurvey}
F.~Nazarieh, Z.~Feng, M.~Awais, W.~Wang, and J.~Kittler, ``A survey of cross-modal visual content generation,'' \emph{IEEE Transactions on Circuits and Systems for Video Technology}, vol.~34, no.~8, pp. 6814--6832, 2024.


\bibitem{spotdiffusion}
S.~Frolov, B.~B.~Moser, and A.~Dengel, ``Spotdiffusion: A fast approach for seamless panorama generation over time,'' in \emph{Proceedings of the
  IEEE/CVF Winter Conference on Applications of Computer Vision}, 2025, pp. 2073--2081.


\bibitem{vae}
D.~P. Kingma and M.~Welling, ``Auto-encoding variational Bayes,'' in
  \emph{Proceedings of the International Conference on Learning Representations}, 2014.


\bibitem{transformer}
A.~Vaswani, N.~Shazeer, N.~Parmar, J.~Uszkoreit, L.~Jones, A.~N.~Gomez, {\L}.~Kaiser, and I.~Polosukhin, ``Attention is all you need,'' \emph{Advances in Neural Information Processing Systems}, vol.~30, 2017.

\bibitem{omnigen}
S.~Xiao, Y.~Wang, J.~Zhou, H.~Yuan, X.~Xing, R.~Yan, C.~Li, S.~Wang,
  T.~Huang, and Z.~Liu, ``Omnigen: Unified image generation,'' in
  \emph{Proceedings of the IEEE/CVF Conference on Computer Vision and Pattern
  Recognition}, 2025, pp.~13294--13304.


\bibitem{mar}
T.~Li, Y.~Tian, H.~Li, M.~Deng, and K.~He, ``Autoregressive image generation without vector quantization,'' \emph{Advances in Neural Information Processing Systems}, vol.~37, pp.~56424--56445, 2024.

\bibitem{nova}
H.~Deng, T.~Pan, H.~Diao, Z.~Luo, Y.~Cui, H.~Lu, S.~Shan, Y.~Qi, and X.~Wang,
  ``Autoregressive video generation without vector quantization,'' in
  \emph{Proceedings of the International Conference on Learning
  Representations}, 2025.


\bibitem{comodgan}
S.~Zhao, J.~Cui, Y.~Sheng, Y.~Dong, X.~Liang, E.~I.-C. Chang, and Y.~Xu,
  ``Large scale image completion via co-modulated generative adversarial
  networks,'' in \emph{Proceedings of the International Conference on Learning
  Representations}, 2021.


\bibitem{unipano}
J.~Ni, C.-B.~Zhang, Q.~Zhang, and J.~Zhang, ``What Makes for Text to 360-degree Panorama Generation with Stable Diffusion?'' \emph{arXiv preprint arXiv:2505.22129}, 2025.


\bibitem{smgd}
X.~Sun, M.~Xu, S.~Li, S.~Ma, X.~Deng, L.~Jiang, and G.~Shen, ``Spherical manifold guided diffusion model for panoramic image generation,'' in \emph{Proceedings of the Computer Vision and Pattern Recognition Conference}, 2025, pp.~5824--5834.


\bibitem{spnd}
X.~Sun, S.~Ma, S.~Li, M.~Xu, J.~Xia, L.~Jiang, X.~Deng, and J.~Wang, ``Spherical-nested diffusion model for panoramic image outpainting,'' in \emph{Proceedings of the Forty-second International Conference on Machine Learning}, 2025.

\bibitem{dreamcube}
Y.~Huang, Y.~Zhou, J.~Wang, K.~Huang, and X.~Liu, ``DreamCube: 3D panorama generation via multi-plane synchronization,'' \emph{arXiv preprint arXiv:2506.17206}, 2025.

\bibitem{latentlabs360}
LatentLabs360, ``Latentlabs360,'' 2023. [Online]. Available: https://civitai.com/models/10753/latentlabs360


\bibitem{interact360}
Z.~Cai, Z.~Huang, X.~Zheng, Y.~Liu, C.~Liu, Z.~Wang, and L.~Wang, ``Interact360: Interactive identity-driven text to 360$^\circ$ panorama generation,'' in \emph{Proceedings of the IEEE Conference on Artificial Intelligence (CAI)}, 2024, pp.~728--736.

\bibitem{virtualreality}
E.~Brivio, S.~Serino, E.~Negro~Cousa, A.~Zini, G.~Riva, and G.~De Leo, ``Virtual reality and 360 panorama technology: A media comparison to study changes in sense of presence, anxiety, and positive emotions,'' \emph{Virtual Reality}, vol.~25, pp.~303--311, 2021.

\bibitem{gaming}
C.-L.~Fan, W.-C.~Lo, Y.-T.~Pai, and C.-H.~Hsu, ``A survey on 360 video streaming: Acquisition, transmission, and display,'' \emph{ACM Computing Surveys}, vol.~52, no.~4, 2019, pp.~1--36.


\bibitem{aisurvey}
H.~Ai, Z.~Cao, and L.~Wang, ``A Survey of Representation Learning, Optimization Strategies, and Applications for Omnidirectional Vision,'' \emph{International Journal of Computer Vision}, vol.~133, no.~8, pp.~4973--5012, 2025.


\bibitem{dreamspace}
B.~Yang, W.~Dong, L.~Ma, W.~Hu, X.~Liu, Z.~Cui, and Y.~Ma, ``Dreamspace: Dreaming your room space with text-driven panoramic texture propagation,'' in \emph{Proceedings of the IEEE Conference on Virtual Reality and 3D User Interfaces (VR)}, 2024, pp.~650--660.


\bibitem{panodreamer}
Z.~Xiong, Z.~Chen, Z.~Li, Y.~Xu, and N.~Jacobs, ``PanoDreamer: Consistent text to 360-degree scene generation,'' in \emph{Proceedings of the Computer Vision and Pattern Recognition Conference}, 2025, pp.~295--304.

\bibitem{immersegen}
J.~Yuan, B.~Yang, K.~Wang, P.~Pan, L.~Ma, X.~Zhang, X.~Liu, Z.~Cui, and Y.~Ma, ``ImmerseGen: Agent-guided immersive world generation with alpha-textured proxies,'' \emph{arXiv preprint arXiv:2506.14315}, 2025.


\bibitem{perf}
G.~Wang, P.~Wang, Z.~Chen, W.~Wang, C.~C. Loy, and Z.~Liu,
  ``Perf: Panoramic neural radiance field from a single panorama,''
  \emph{IEEE Transactions on Pattern Analysis and Machine Intelligence},
  vol.~46, no.~10, pp.~6905--6918, 2024.


\bibitem{360dvd}
Q.~Wang, W.~Li, C.~Mou, X.~Cheng, and J.~Zhang, ``360dvd: Controllable panorama video generation with 360-degree video diffusion model,'' in \emph{Proceedings of the IEEE/CVF Conference on Computer Vision and Pattern Recognition}, 2024, pp.~6913--6923.


\bibitem{dynamicscaler}
J.~Liu, S.~Lin, Y.~Li, and M.-H.~Yang, ``Dynamicscaler: Seamless and scalable video generation for panoramic scenes,'' in \emph{Proceedings of the Computer Vision and Pattern Recognition Conference}, 2025, pp.~6144--6153.


\bibitem{panodit}
M.~Zhang, Y.~Chen, R.~Xu, C.~Wang, J.~Yang, W.~Meng, J.~Guo, H.~Zhao, and X.~Zhang, ``PanoDit: Panoramic videos generation with diffusion transformer,'' in \emph{Proceedings of the AAAI Conference on Artificial Intelligence}, vol.~39, no.~10, 2025, pp.~10040--10048.

\bibitem{videopanda}
K.~Xie, A.~Sabour, J.~Huang, D.~Paschalidou, G.~Klar, U.~Iqbal, S.~Fidler, and X.~Zeng, ``VideoPanda: Video panoramic diffusion with multi-view attention,'' \emph{arXiv preprint arXiv:2504.11389}, 2025.

\bibitem{panowan}
Y.~Xia, S.~Weng, S.~Yang, J.~Liu, C.~Zhu, M.~Teng, Z.~Jia, H.~Jiang, and B.~Shi, ``PanoWan: Lifting diffusion video generation models to 360-degree with latitude/longitude-aware mechanisms,'' \emph{arXiv preprint arXiv:2505.22016}, 2025.


\bibitem{t2v1}
J.~An, S.~Zhang, H.~Yang, S.~Gupta, J.-B.~Huang, J.~Luo, and X.~Yin, ``Latent-shift: Latent diffusion with temporal shift for efficient text-to-video generation,'' \emph{arXiv preprint arXiv:2304.08477}, 2023.


\bibitem{t2v2}
Y.~Guo, C.~Yang, A.~Rao, Z.~Liang, Y.~Wang, Y.~Qiao, M.~Agrawala, D.~Lin,
  and B.~Dai, ``AnimateDiff: Animate your personalized text-to-image diffusion
  models without specific tuning,'' in \emph{Proceedings of the International
  Conference on Learning Representations}, 2024.


\bibitem{t2v3}
J.~Xing, M.~Xia, Y.~Liu, Y.~Zhang, Y.~Zhang, Y.~He, H.~Liu, H.~Chen, X.~Cun, X.~Wang, \emph{et al.}, ``Make-your-video: Customized video generation using textual and structural guidance,'' \emph{IEEE Transactions on Visualization and Computer Graphics}, 2024.


\bibitem{t2v4}
H.~Wang, C.-Y.~Ma, Y.-C.~Liu, J.~Hou, T.~Xu, J.~Wang, F.~Juefei-Xu, Y.~Luo, P.~Zhang, T.~Hou, \emph{et al.}, ``Lingen: Towards high-resolution minute-length text-to-video generation with linear computational complexity,'' in \emph{Proceedings of the Computer Vision and Pattern Recognition Conference}, 2025, pp.~2578--2588.

\bibitem{dit}
W.~Peebles and S.~Xie, ``Scalable diffusion models with transformers,'' in \emph{Proceedings of the IEEE/CVF International Conference on Computer Vision}, 2023, pp.~4195--4205.

\bibitem{wan}
Team~Wan, A.~Wang, B.~Ai, B.~Wen, C.~Mao, C.-W.~Xie, D.~Chen, F.~Yu, H.~Zhao, J.~Yang, \emph{et al.}, ``Wan: Open and advanced large-scale video generative models,'' \emph{arXiv preprint arXiv:2503.20314}, 2025.


\bibitem{3dscenesurvey}
B.~Wen, H.~Xie, Z.~Chen, F.~Hong, and Z.~Liu,
  ``3D scene generation: A survey,''
  \emph{arXiv preprint arXiv:2505.05474}, 2025.


\bibitem{imagine360}
J.~Tan, S.~Yang, T.~Wu, J.~He, Y.~Guo, Z.~Liu, and D.~Lin,
``Imagine360: Immersive 360 video generation from perspective anchor,''
\emph{arXiv preprint arXiv:2412.03552}, 2024.


\bibitem{360-1m}
M.~Wallingford, A.~Bhattad, A.~Kusupati, V.~Ramanujan, M.~Deitke,
A.~Kembhavi, R.~Mottaghi, W.-C.~Ma, and A.~Farhadi,
``From an image to a scene: Learning to imagine the world from a million 360 videos,''
\emph{Advances in Neural Information Processing Systems}, vol.~37, pp.~17743--17760, 2024.

\bibitem{odv360}
M.~Cao, C.~Mou, F.~Yu, X.~Wang, Y.~Zheng, J.~Zhang, C.~Dong,
G.~Li, Y.~Shan, R.~Timofte, \emph{et al.},
``NTIRE 2023 Challenge on 360° Omnidirectional Image and Video Super-Resolution: Datasets, Methods and Results,''
\emph{Proceedings of the 2023 IEEE/CVF Conference on Computer Vision and Pattern Recognition Workshops (CVPRW)}, 
pp.~1731--1745, 2023.


\bibitem{videollama2}
Z.~Cheng, S.~Leng, H.~Zhang, Y.~Xin, X.~Li, G.~Chen, Y.~Zhu,
W.~Zhang, Z.~Luo, D.~Zhao, \emph{et al.},
``Videollama 2: Advancing spatial-temporal modeling and audio understanding in video-LLMs,''
\emph{arXiv preprint arXiv:2406.07476}, 2024.

\bibitem{qwen}
S.~Bai, K.~Chen, X.~Liu, J.~Wang, W.~Ge, S.~Song, K.~Dang,
P.~Wang, S.~Wang, J.~Tang, \emph{et al.},
``Qwen2.5-VL Technical Report,''
\emph{arXiv preprint arXiv:2502.13923}, 2025.

\end{thebibliography}
\end{document}